\documentclass[manuscript,screen,review=false]{acmart}

\makeatletter
\@ifundefined{Bbbk}{}{}
\makeatother

\usepackage{graphicx}%
\usepackage{multirow}%
\usepackage{amsmath,amssymb}%
\usepackage{amsthm}%
\usepackage{mathrsfs}%
\usepackage[title]{appendix}%
\usepackage{xcolor}%
\usepackage{textcomp}%
\usepackage{manyfoot}%
\usepackage{booktabs}%
\usepackage{algorithm}%
\usepackage{algorithmicx}%
\usepackage{algpseudocode}%
\usepackage{listings}%
\usepackage{rotating}

\makeatletter
\@ifundefined{Bbbk}{
  \usepackage{amssymb}
}{}
\makeatother
\usepackage{textcomp}
\usepackage{multirow}
\usepackage{float}
\usepackage{makecell}
\usepackage{stfloats}
\usepackage{hyperref}
\usepackage{url}
\usepackage{pifont}
\usepackage{enumitem}

\usepackage{booktabs}    
\usepackage{ragged2e}    
\usepackage{forest}
\usepackage{tabularx}
\usepackage{tikz-qtree}
\usetikzlibrary{patterns,arrows,positioning,calc,fadings,shapes,decorations.markings}
\usepackage{xcolor}
\definecolor{Red}{RGB}{190,20,42}
\definecolor{CY}{RGB}{233,246,254}
\theoremstyle{thmstyleone}%
\theoremstyle{thmstyletwo}%

\theoremstyle{thmstylethree}%
\AtBeginDocument{%
  }

\setcopyright{acmlicensed}
\copyrightyear{2026}
\acmYear{2026}
\acmConference[XX]{ACM TKDD}{June 03--05,
  2026}{Woodstock, NY}
\acmISBN{978-1-4503-XXXX-X/2026/10}

\begin{document}

\title{Data Optimization for LLMs: A Survey}

\author{Ou Wu}
\authornote{Corresponding author.}
\thanks{This study is supported by NSFC No. 42550187 and 62476191.}
\email{wuou@ucas.ac.cn}
\affiliation{%
  \institution{Hangzhou Institute for Advanced Study, University of Chinese Academy of Sciences}
  \city{Hangzhou}
  \country{China}
}


\begin{abstract}
Large language models (LLMs) have attracted great attention in both academic and industrial communities in recent years. Numerous survey studies have covered various aspects of LLMs, such as learning strategies, applications, alignment, explainability, and evaluation. However, little headway has been made in reviewing data-related aspects for LLMs, despite data being recognized as a key factor contributing to the emergence of the LLMs era. A wide range of data optimization algorithms and techniques are employed in LLMs' training and application processes. These techniques often originate from diverse deep learning domains, with their theoretical inspirations and heuristic motivations appearing unrelated to each other. This study aims to develop a comprehensive framework for LLMs' data optimization techniques to enhance data utilization and efficiency. We first discuss data challenges in LLMs' training and inference. Second, we summarize data perception dimensions for evaluating LLMs' data properties. These dimensions primarily include quality, difficulty, diversity, and redundancy, laying a foundation for subsequent optimization. Third, existing optimization techniques are reviewed and categorized into four groups: data selection, augmentation, reweighting, and others. Fourth, we analyze interconnections between these categories. Finally, we identify key challenges and future directions for LLMs' data optimization compared with conventional deep learning approaches.
\end{abstract}



\received[Accepted]{20 Sep. 2026}

\maketitle

\section{Introduction}\label{sec:introduction}
LLMs have gained significant attention in the fields of deep learning and broader artificial intelligence research in recent years. Owing to their remarkable performance, these models have also garnered substantial interest from governments and the general public. Some researchers even suggest that LLMs exhibit strong potential for Artificial General Intelligence~\cite{10697304}. The success of LLMs can be attributed to three main factors: (1) advanced DNN architectures and learning algorithms, particularly the Transformer~\cite{vaswani2017attention}, (2) ultra-large-scale, high-quality training datasets~\cite{kaplan2020scaling,gao2020pile}, and (3) efficient computational infrastructure, such as GPUs~\cite{narayanan2021efficient}. Thus, high-quality data is indispensable for LLM development. Recent evidence suggests that exhaustion of publicly available datasets will bottleneck large-scale model performance~\cite{villalobos2022data}, further highlighting data's critical role.

Researchers have devoted substantial efforts to improving the training and inference performance of LLMs from a data-centric perspective. Based on existing literature, data optimization for LLMs primarily addresses the following data defects~(as shown in Fig.~\ref{fig1}):
    \begin{figure*}[t] 
    \centering \includegraphics[width=0.860\linewidth,height=1.42in]{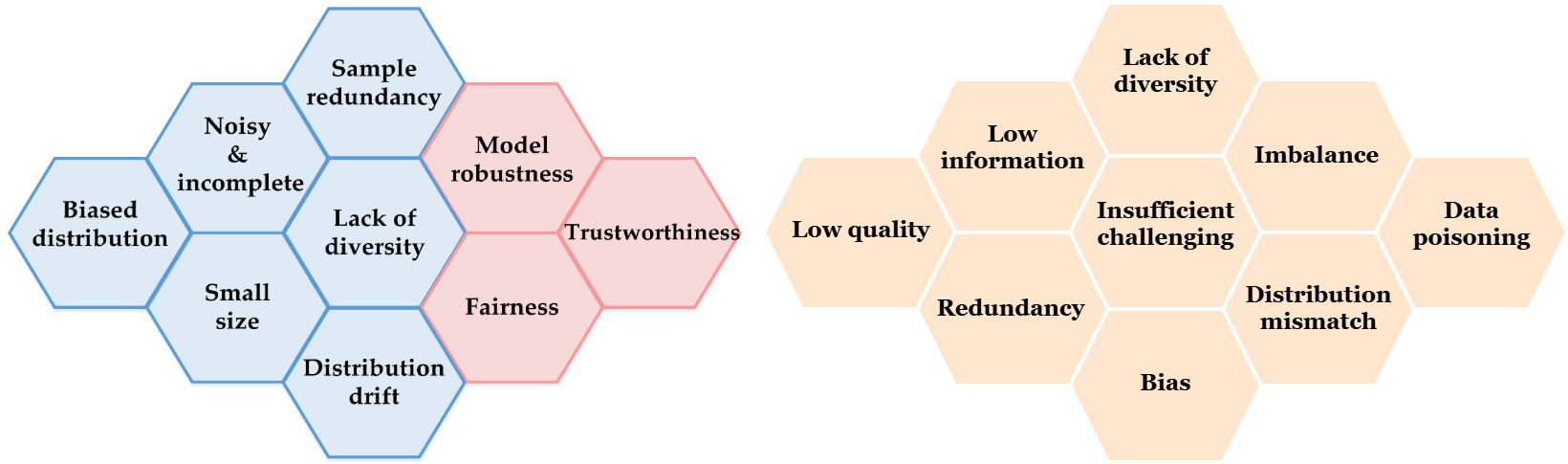}    \vspace{-0.0in}
    \caption{Nine issues around the data in conventional deep learning~(left)~\cite{wu2024data}, and around the data in LLMs~(right). Key mappings: (1) ``Distribution drift'' (conventional DL) corresponds to ``Distribution mismatch'' (LLMs)—with LLM-specific extensions like inference-time prompt distribution shift; (2) ``Noisy'' (conventional DL) aligns with ``Low quality'' (LLMs)—including LLM-unique formatting chaos; (3) {``Biased distribution''} (conventional DL) maps to ``Imbalance'' (LLMs)—including novel multi-domain mixture proportion bias. Remaining issues (e.g., data poisoning) are LLM-specific or extended from conventional DL.}
    \label{fig1}
     \vspace{-0.0in}
\end{figure*}
\begin{itemize}
    \item \textbf{Low quality}: The pretraining of LLMs is fundamentally constrained by two critical data quality issues that pervade their training corpora. The first is the prevalence of noise and errors. Web-scale training data inherently contains significant noise, including factual inaccuracies, grammatical errors, and semantically inconsistent statements. The second is the structural inconsistencies and formatting chaos in the original training data. The latter can be effectively addressed by data cleaning techniques, which are not the focus of this study.

    \item \textbf{Low information}: This defect mainly refers to domain-specific token density and semantic coverage and  lower values mean insufficient domain knowledge per unit text. The supervised fine-tuning~(SFT) and inference stages of LLMs usually suffer from the lack of sufficient information. Many vertical AI applications~(e.g., biomedical diagnostics, legal analysis) face acute shortages of domain-specific training data. The SciBERT dataset, for instance, contains merely 12\% of the token density found in general web text~\cite{beltagy2019scibert}, forcing models to rely on inadequate data resources. On the other hand, inputs (e.g., instructions, demonstrations) in inference usually contain limited information, hindering high-quality responses for LLMs.

    \item \textbf{Redundancy}: While LLMs are typically trained on massive datasets, not all data contributes equally to model performance. Two distinct but interrelated redundancy problems emerge in LLM development:  
    1) Web-crawled training corpora exhibit significant duplication, with studies showing 3--14\% of tokens in datasets like Common Crawl appearing in near-identical form across multiple documents~\cite{lee2022deduplicating}; 2) Even with deduplication, not all samples are necessary in training.

    \item \textbf{Lack of diversity}: LLMs face significant challenges due to the lack of diversity in their training corpora, which manifests through the disproportionate representation of certain linguistic patterns, styles, or domain-specific knowledge. While web-scale datasets may appear comprehensive, they often exhibit systemic biases in which particular writing styles~(e.g., Wikipedia's encyclopedic tone) and topical coverage become disproportionately prevalent. This homogeneity can lead to spurious correlations between specific languages, styles, or semantic topics. Crucially, while this issue shares similarities with data redundancy in that both involve skewed distributions, the lack of diversity is fundamentally distinct—it pertains to the absence of meaningful variation rather than mere duplication of content.

    \item \textbf{Insufficiently challenging}: A critical yet understudied limitation in SFT for LLMs lies in the prevalence of insufficiently challenging data, which manifests through two interrelated dimensions: inadequate data difficulty and low data complexity. Over-reliance on simplistic training data~(e.g., short sentences) leads to premature model convergence and weak generalization, while the lack of complex examples~(e.g., abbreviated Chain-of-Thought sequences) inherently restricts systematic reasoning capabilities.

    \item \textbf{Bias}: This issue shares some commonalities with the lack of diversity. The pretraining of LLMs inherently amplifies social biases embedded in their training corpora, manifesting through distorted representations across dimensions such as demographics and geographic perspectives. Web-scale datasets systematically overrepresent majority perspectives. In addition, gender stereotypes are particularly pervasive: LLMs exhibit a 3–6 times higher propensity to select occupations aligned with gender stereotypes~\cite{Kotek_2023}.

\item \textbf{Imbalance:} This issue also heavily constrains the development of LLMs across both pretraining and SFT phases. During pretraining, web-scale datasets exhibit severe domain imbalance, where technical content and low-resource languages are dramatically underrepresented compared to high-frequency web genres, forcing models to develop uneven world knowledge that privileges dominant contents and semantic patterns. The instruction-tuning datasets often overrepresent popular tasks while neglecting complex reasoning formats, creating artificial capability ceilings as models overfit to prevalent task structures.

\item \textbf{Distribution mismatch:} LLMs face a persistent distribution mismatch issue throughout their lifecycle, beginning with pretraining and extending to inference-time deployment and continual learning scenarios. Temporal drift in pretraining causes models to struggle with evolving language and facts. During inference, deviations from training data often generate poor responses. Continual learning further worsens performance through catastrophic forgetting during domain adaptation. Notably, LLM-specific inference-time prompt distribution shift~(e.g., few-shot exemplar ordering impacting outputs) does not exist in conventional deep learning.

\item \textbf{Data poisoning:} The growing threat of data poisoning in LLM training has emerged as a critical security concern, particularly given the reliance on web-scale corpora with minimal curation. Maliciously crafted training samples can induce systematic vulnerabilities, enabling targeted misinformation propagation, backdoor attacks, or subtle behavioral manipulation that evades conventional anomaly detection~\cite{goldblum2022dataset}. A recent study~\cite{Alber2025medical} demonstrates that as little as 0.001\% poisoned data can reliably compromise model outputs while maintaining standard benchmark performance, highlighting the stealthiness of such attacks.
\end{itemize}

Wu and Yao~\cite{wu2024data} briefly summarized data-related issues concerning conventional deep learning scenarios as shown in Fig.~1(left). Three key distinctions drive differences in data focus between LLMs and conventional models: (1) LLMs require holistic optimization across pretraining, SFT, and inference stages versus conventional models' training-centric approach; (2) general-purpose LLMs contrast with application-specific conventional tasks; (3) LLMs must address safety/value alignment challenges beyond conventional accuracy prioritization.

The above-mentioned issues for LLMs mainly refer to data defects. There are other issues pertaining to data utility:
\begin{itemize}
\item \textbf{Data order:} The sequential organization of data during both training and inference phases constitutes a critical yet underappreciated determinant for LLMs. In SFT, curriculum learning strategies that progressively escalate task complexity have been shown to improve final performance compared to random ordering~\cite{lee2024ITW}. During inference, the ordering of few-shot exemplars significantly impacts output quality~\cite{lu2022fantastically}.
\item \textbf{Mixture proportions:} The pretraining for LLMs involves multi-domain datasets. Empirical studies reveal that the setting of mixture proportions for multiple data domains has emerged as a critical determinant of model performance~\cite{brown2020}. This sensitivity necessitates systematic balancing between domain diversity and task relevance, particularly given the varying linguistic patterns, conceptual densities, and noise levels inherent in heterogeneous corpora. This sensitivity is LLM-exclusive as conventional DNN models usually train on single-domain data and require no multi-domain proportion tuning, whereas LLM pretraining merges general and domain-specific corpora.
\end{itemize}

{Crucially, each of these data defects and utility considerations directly impacts specific aspects of LLM behavior. Low-quality data undermines trustworthiness by causing factual hallucinations; insufficiently challenging data limits effectiveness by preventing the development of complex reasoning; lack of diversity harms fairness by amplifying societal bias; and data redundancy impairs efficiency by wasting computational resources. Data optimization techniques investigated in this survey serve as the primary means to address these issues, thereby enhancing the overall capabilities of LLMs. Throughout this paper, we will explicitly connect each optimization method to the LLM aspects it influences, providing a holistic view of how data-centric interventions shape model behavior.}

To address the above defects and improve data utility for LLMs, numerous new methodologies and various empirical studies have been proposed in previous literature. {Similar to the survey study summarized by Wu and Yao~\cite{wu2024data}, this paper focuses on methods that improve LLM performance by intervening on the \emph{data} used in training or inference, rather than by designing new model architectures or learning algorithms. We refer to this family of methods as \textbf{data optimization for LLMs}. More specifically, data optimization denotes the systematic improvement of data usage through operations such as selecting, augmenting, reweighting, re-ordering, or compressing data, with the goal of improving one or more downstream objectives, including effectiveness, efficiency, fairness, and trustworthiness.}

Existing LLM studies addressing data deficiency and utility usually originate from disparate subfields with divergent motivations and focuses. A systematic synthesis and interdisciplinary integration of these investigations could facilitate mutual understanding among different data optimization paths, foster deeper comprehension of data's multifaceted roles in LLM development, and ultimately inspire novel approaches to data optimization methodologies for LLMs. Such an endeavor would not only promote the training and inference of LLMs but also establish foundational principles for data-centric LLM research paradigms.

The differences between our survey and existing surveys on LLMs in relevant areas—including data selection, data augmentation, adversarial training, and dataset distillation—lie in two aspects. First, this survey takes a much broader view of previous data-centric studies concerning LLMs. Consequently, a more comprehensive taxonomy is constructed, where most relevant surveys only address a subdivision of our taxonomy. Second, connections among different optimization paths are discussed in this survey, which are typically not addressed in existing surveys. Our contributions are summarized as follows:
 \begin{itemize}
\item Methodologies to address distinct data defects and to improve data utility in LLM pretraining and inference are organized within a new taxonomy. To our knowledge, this is the first work that aims to construct a data-centric taxonomy focusing on data optimization across multiple subdivisions for LLMs.
\item Several fundamental dimensions are summarized for the analysis of both training and inference data for LLMs. A new taxonomy is also constructed for these analytical dimensions.

\item Connections among different data optimization techniques are established. Additionally, potential future research directions are outlined.
 \end{itemize}

This paper is organized as follows: Section~\ref{sec2} reviews existing surveys; Section~\ref{sec3} establishes data optimization overview; Sections~\ref{sec4}-\ref{sec5} detail data perception and optimization techniques respectively; Section~\ref{sec6} analyzes technical interconnections and proposes future directions; Section~\ref{sec7} concludes the study.

\begin{table*}
\centering
\small
\caption{Several relevant data-centric survey studies~(most are about LLMs).}\vspace{-0.08in}
\renewcommand{\arraystretch}{1.6} 
\begin{tabularx}{\textwidth}{| *{5}{>{\centering\arraybackslash}X|} } 
\hline
\textbf{Survey studies} & \textbf{Topics} & \textbf{Techniques} & \textbf{Stages} &  \textbf{Perception} \\ 
\hline\hline
Albalak et al.~\cite{albalak2024}, Wang et al.~\cite{wang2024surveyd} & Data selection & Selection, re-ordering & Training, Inference & Mentioned but not systematically elaborated \\ 
\hline
Ding et al.~\cite{ding2024data}, Chai et al.~\cite{chai2025text} & Data augmentation & Augmentation, synthesis & Training, Inference & Not referred \\ 
\hline
Wang et al.~\cite{wang2024survey} & Synthesis, augmentation & Synthesis, augmentation & Training, Inference & Mentioned but not systematically elaborated\\ 
\hline
Das et al.~\cite{10.1145/3712001}, Zhao et al.~\cite{zhao2025dat} & Security, poisoning & Selection & Training, Inference & Not referred \\ 
\hline
Navigli et al.~\cite{10.1145/3597307}, Chu et al.~\cite{10.1145/3682112.3682117} & Bias and fairness & Selection, augmentation, reweighting & Training, Inference & Data balance\\ 
\hline
Ferdaus et al.~\cite{ferdaus2024trust}, Huang et al.~\cite{huang2023sur}& Trustworthiness & Selection, augmentation, reweighting & Training, Inference & Data quality, balance \\ 
\hline
Li et al.~\cite{li2024goal}, Li et al.~\cite{li2024pro} & Prompt engineering & Selection, prompt compression & Inference & Not referred \\ 
\hline
Wang et al.~\cite{wang2024dat} & Data management & Selection & Training & Quality, difficulty, diversity, redundancy\\ 
\hline
Wu and Yao~\cite{wu2024data} & Data optimization~(for deep learning)  & Selection, augmentation, reweighting, distillation & Training & Distribution, cleanliness, difficulty, uncertainty, balance, consistency, value \\
\hline
Our survey & Data optimization~(for LLMs) & Selection, augmentation, reweighting, re\-ordering, distillation, compression & Training, Inference & Quality, difficulty, diversity, uncertainty, value, redundancy, balance, contamination, coverage \\ 
\hline
\end{tabularx}
 \label{tbl1}
\end{table*}

\section{Related Studies}\label{sec2}

The data defects and utilities mentioned earlier have led to many specialized research areas in LLMs. Researchers have conducted numerous surveys to summarize progress in these fields. This section highlights key survey papers related to the mentioned data defects and utilities for LLMs.

\textbf{Data quality for LLMs}: The construction of an LLM is highly sensitive to the quality of the training data. There is no consensus definition for data quality. For example, in some studies~\cite{miranda2024}, diversity is leveraged as an indicator for data quality, while in other studies~(e.g.,~\cite{havrilla2024survey}), diversity is considered as an independent measure for training data. In nearly all previous studies, data quality typically refers to correctness or cleanness. Yu et al.~\cite{10.1145/3691620.3695061} conducted a survey of 219 LLM practitioners from 23 countries to identify important characteristics in terms of data quality. In our survey, data quality is listed as the primary data perception dimension.

\textbf{Data selection for LLMs}: Data selection receives considerable attention in the LLM era, as it can improve both training performance and efficiency. Albalak et al.~\cite{albalak2024} conducted a comprehensive literature review of existing data selection studies according to employed utility functions and selection strategies. Data selection is considered an indispensable data optimization technique in our survey.

\textbf{Data augmentation for LLMs}: This technique is widely applied in both the training and inference phases of LLMs~\cite{ding2024data}. Chai et al.~\cite{chai2025text} divided existing data augmentation methods for LLMs into four categories: simple, prompt-based, retrieval-based, and hybrid augmentation. Our survey treats data augmentation as an important data optimization technique and provides an alternative classification.

\textbf{Data synthesis for LLMs}: This technique is similar to data augmentation as it also aims to expand the training set by creating new data. Wang et al.~\cite{wang2024survey} grouped data synthesis and augmentation into a single paradigm called data generation. In this study, pure data synthesis is not considered a data optimization technique because our definition of data optimization is limited to techniques that optimize existing training data. However, the augmentation techniques discussed herein can be applied to post-process synthetic data to improve its utility.

\textbf{Data security for LLMs}: Recent survey studies have systematically examined data security challenges in LLMs, highlighting critical threats such as data poisoning, adversarial attacks, and privacy leakage during both pretraining and SFT~\cite{10.1145/3712001}. LLMs are particularly vulnerable to backdoor attacks and membership inference due to their reliance on massive, multi-source training datasets, where even minimal poisoned samples can significantly degrade model integrity.

\textbf{Bias and fairness in LLMs}: The widespread deployment of LLMs has brought increasing attention to their inherent biases and the consequential fairness issues, both of which predominantly stem from imperfections in the training data. Navigli et al.~\cite{10.1145/3597307} discussed data selection bias in the construction of training corpora and summarized the social bias evidenced in text generated by LLMs trained on these corpora. Chu et al.~\cite{10.1145/3682112.3682117} presented a comprehensive overview of recent literature concerning fair LLMs.

\textbf{Trustworthy LLMs}: As LLMs increasingly influence domains such as healthcare, education, and decision-making, ensuring their trustworthiness has become a multidisciplinary imperative. Trustworthy LLMs must also address hallucinations—factually flawed or ungrounded outputs. Techniques such as data curation remove noise, misinformation, or toxic content from datasets. Synthetic data generation further enables training on privacy-sensitive domains without compromising individual anonymity. Ferdaus et al.~\cite{ferdaus2024trust} established a new framework to assess trust and analyzed the complex trust dynamics of LLMs.

\textbf{Prompt engineering}: This data-centric technique systematically designs input prompts to steer LLMs toward desired outputs without altering underlying architectures. By refining textual cues—such as explicit instructions or structured templates—it mitigates ambiguities in model interpretation, enhances zero-shot/few-shot learning, and reduces hallucination risks~\cite{li2024goal}.

{\textbf{Cross-domain methodological transfers}: Data optimization principles developed for tabular and structured data---such as RL-based data transformation, generative data repairing, and feature-engineering automation---are increasingly relevant to LLM data optimization, particularly as LLMs are deployed over structured knowledge bases and tool-augmented environments. 
Recent surveys on tabular data-centric AI provide methodological bridges that can inspire LLM training corpus optimization, where tabular metadata (e.g., document provenance, quality scores, temporal stamps) can be optimized jointly with raw text~\cite{wang2026toward}.}

There are also other surveys related to this study. Liu et al.~\cite{liu2024datasets} conducted a detailed review of public LLM datasets and categorized them into five types: pretraining corpora, instruction fine-tuning datasets, preference datasets, evaluation datasets, and traditional NLP datasets. Most methods listed in this survey are performed on these datasets. Data contamination is also an important issue for LLM training~\cite{xu2024benchmark}. Some survey studies focus on specific types of data properties or metrics such as data diversity~\cite{havrilla2024survey}, uncertainty~\cite{shorinwa2024survey}, and similarity~\cite{10584318}. Table~\ref{tbl1} lists several recent data-centric survey studies about LLMs. Our survey differs from these reviews in the following aspects:
\begin{itemize}
\item[(1)] Our study reviews a wide range of data optimization techniques related to both data defects and utilities for LLMs, whereas previous surveys focus on either a single data aspect or an independent optimization path.
\item[(2)] Data optimization highly depends on an explicit or implicit prerequisite step—data perception of the involved data. Previous surveys typically ignore this step, while our study introduces it in detail.
\item[(3)] We discuss connections among different optimization technical paths, which previous surveys limited to single aspects or paths do not address.
\end{itemize}

\begin{figure}[t] 
    \centering \includegraphics[width=0.616\linewidth,height=1.86in]{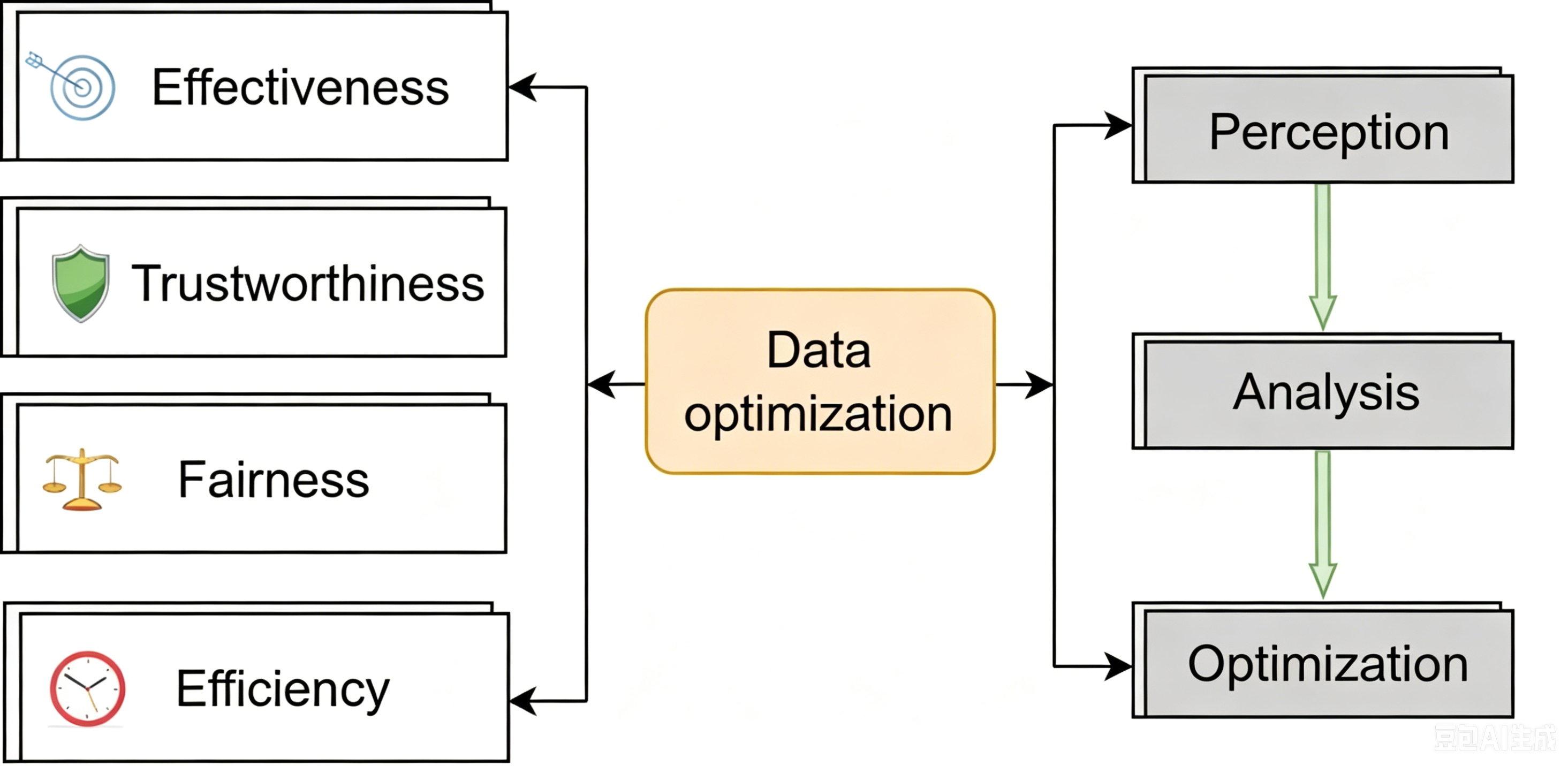}    \vspace{-0.0in}
    \caption{{The main goals~(left) and pipeline of data optimization for LLMs~(Data synthesis, when performed without corpus anchoring, is excluded from this pipeline; when used to augment existing data, it is subsumed under data augmentation.)}}
    \label{fig2}
     \vspace{-0.1in}
\end{figure}

\section{Overview of Data Optimization Studies for LLMs}\label{sec3}

 To avoid creating an overly large taxonomy as established for data optimization in conventional deep learning~\cite{wu2024data}, this section presents an overview of data optimization studies for LLMs.

\subsection{Optimization Goals}
Fig.~\ref{fig2} illustrates the four main optimization goals: effectiveness, trustworthiness, fairness, and efficiency. {As introduced in Section~\ref{sec:introduction}, the data defects and utility-related challenges of LLMs motivate the need for data optimization. The four optimization goals discussed in this section are not a separate taxonomy, but rather the high-level LLM objectives that data optimization seeks to improve in response to those data-side issues. The framework in this section thus connects the causes identified from the data side with the outcomes expected on the model side.} Compared with conventional data optimization goals, generalization has been replaced by effectiveness and trustworthiness. This variation stems from two key reasons. First, data optimization in LLMs applies to both training and inference stages, where effectiveness encompasses generalization during training and accuracy during inference. Second, trustworthiness represents a fundamental prerequisite for LLMs to provide reliable services.

\begin{itemize}
\item \textbf{Effectiveness:} This goal comprises two aspects. The first aspect is generalization, which remains a core concept in machine learning and includes representation capability for input data. The second aspect focuses on inference accuracy for the trained LLM's outputs.

\item \textbf{Trustworthiness:} This goal has emerged as a paramount concern in current LLM research, positioning both algorithmic and data optimization as critical pathways to address these challenges. It encompasses three main components: (1) hallucination mitigation, (2) human-value alignment, and (3) model robustness (including adversarial robustness, out-of-distribution robustness, etc.).

\item \textbf{Fairness:} {This goal represents another primary objective in algorithmic and data optimization, addressing biases inherent in training data and model behavior for LLMs. Techniques such as debiasing algorithms, fairness-aware loss functions, and data reweighting aim to reduce demographic disparities in model outputs. However, these interventions may also introduce trade-offs with performance or other attributes~\cite{anthis2025impossibility}, so fairness should ideally be considered jointly with other optimization goals.}

\item \textbf{Efficiency:} This objective holds significant practical importance as real-world learning tasks frequently face critical computational and storage constraints. Developing optimal strategies to eliminate redundant training samples while preserving data diversity and informational value requires systematic exploration. Strategic data pruning not only reduces time complexity but also addresses storage limitations, thereby enhancing scalability without compromising model performance.
\end{itemize}

{While the four optimization goals are presented separately for analytical clarity, they are interdependent in practice, exhibiting both synergies and inherent tensions. Several goals naturally reinforce each other. Improving data quality and diversity, which primarily targets effectiveness, often contributes to reducing model biases and thereby promotes fairness. Similarly, filtering toxic or harmful content, a key operation for trustworthiness, simultaneously serves the purpose of building safer and more aligned models. These synergistic relationships suggest that well-designed data optimization can potentially advance multiple objectives at once.}

{However, pursuing one goal may also come at the expense of another. A representative example is the fairness--accuracy trade-off: interventions that reduce demographic disparities, such as data reweighting or debiasing-oriented sampling, may also lower overall performance or shift errors across groups~\cite{anthis2025impossibility}. Related tensions have also been observed between fairness and privacy. Fine-tuning LLMs to enhance privacy awareness has been shown to inadvertently reduce fairness sensitivity~\cite{qian-etal-2025-tug}, while differential privacy mechanisms may disproportionately affect minority groups through the injected noise~\cite{xu2026differential}. Qian et al.~\cite{qian-etal-2025-tug} trace this to neuron 
semantic superposition; Karvonen and Marks~\cite{karvonen2025robustly} 
show that realistic context induces 12\% disparities; and Li 
et al.~\cite{li2024survey} note that intrinsic and extrinsic metrics 
are not necessarily correlated. In addition, interventions targeting a single attribute such as gender 
may unintentionally worsen biases on other untargeted attributes, 
highlighting the need for intersectional evaluation~\citep{paraschou2026intra}. Selective inference-time prompting offers a lightweight alternative, improving fairness without retraining costs~\cite{ren2026whentoinvoke}.}

\begin{figure}[t] 
    \centering \includegraphics[width=0.76\linewidth,height=2.16in]{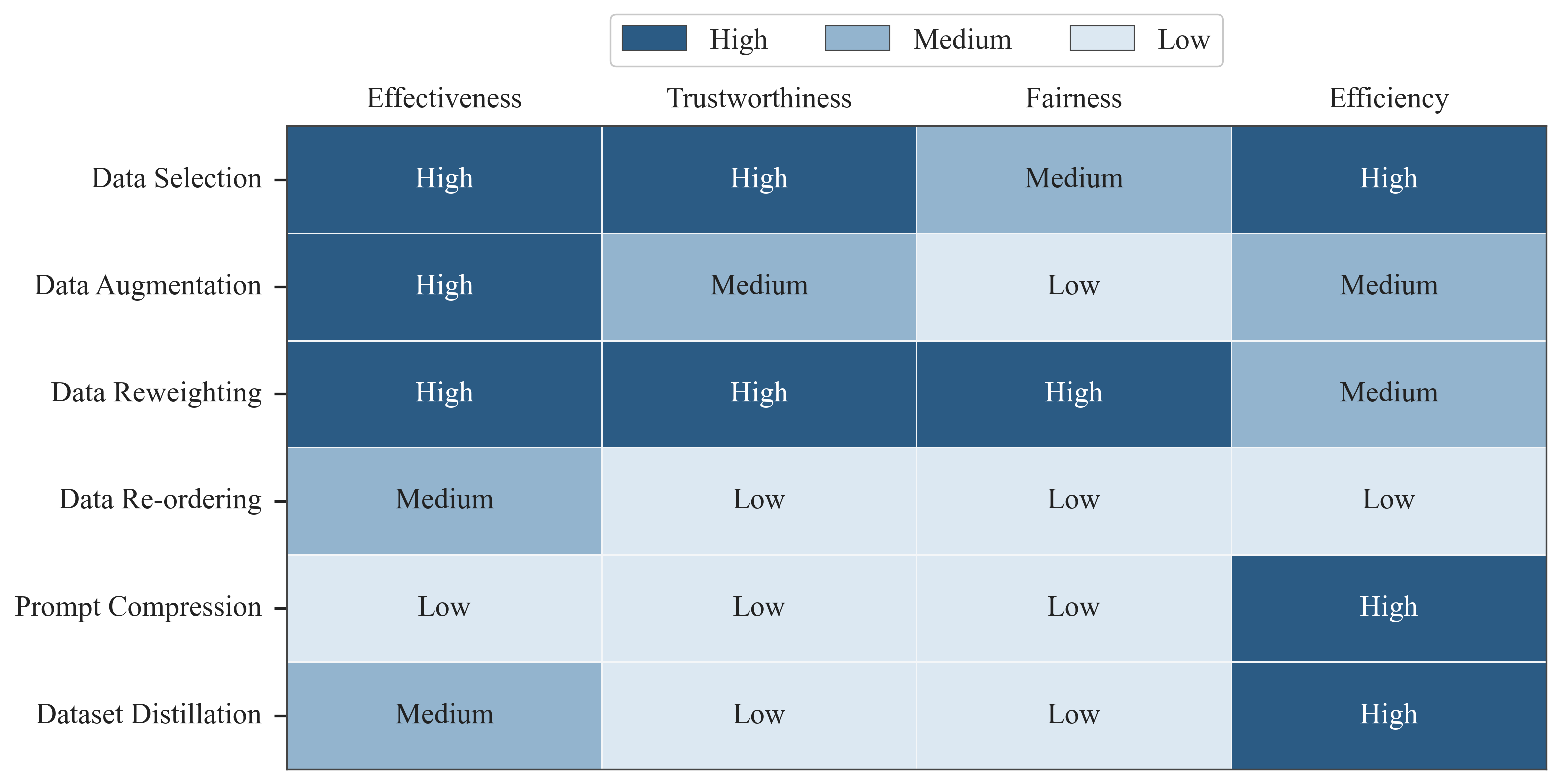}    \vspace{-0.0in}
    \caption{{Impact intensity of data optimization techniques on key LLM aspects. \textbf{Reading guide:} Rows represent six technical categories; columns represent four LLM aspects. Dark blue (High) indicates the primary target of the technique; medium blue (Medium) indicates significant secondary effects; light blue (Low) indicates minor or indirect effects. For example, Data Selection is primarily designed for Effectiveness and Efficiency, yet attains High trustworthiness because quality filtering removes noisy samples that cause hallucinations.}}
    \label{fig2-1}
     \vspace{-0.1in}
\end{figure}

{These observations carry practical implications: in LLM data optimization, no single goal should ideally be pursued in isolation. Improvements along one dimension may come with costs along others, and the preferred balance depends on the application context. Consequently, evaluation demands multi-dimensional benchmarks that holistically assess impact across all relevant goals, rather than focusing on a single metric like downstream accuracy. Fig.~\ref{fig2-1} maps specific data optimization techniques (to be detailed in Section~\ref{sec5}) to these four goals, based on our 
literature survey, revealing that most techniques impact multiple aspects simultaneously. For instance, data selection primarily targets effectiveness and efficiency but also contributes to trustworthiness by filtering noisy samples. This multi-aspect impact underscores the need for deliberate trade-off management when composing optimization techniques (see Section~\ref{sec5.6}).}

\subsection{Optimization Pipeline}
The pipeline of data optimization in LLMs follows a similar structure to conventional deep learning, consisting of three main steps: data {perception}, data analysis, and data optimization. Although many studies do not explicitly include all three steps~(e.g., an explicit perception step is omitted in many studies), these omissions are typically compensated for by prior knowledge or implicit rational assumptions. {We note that the term \emph{data optimization} operates at two levels in this survey: 
it denotes the overarching research domain (the title-level topic), 
and also the third stage of the pipeline where concrete interventions (selection, augmentation, reweighting, etc.) are applied. 
The intended meaning is always clear from context.}

{In this framework, \emph{data perception} characterizes data properties, \emph{data analysis} interprets these properties in terms of defects and utility, and \emph{data optimization} uses the resulting signals to guide interventions on the data pipeline:}
\begin{itemize}
\item \textbf{{Data perception}:} {This initial step refers to the process of {measuring or characterizing the properties of training or inference data} that may affect model behavior. Its goal is to identify intrinsic data characteristics---such as quality, difficulty, diversity, uncertainty, value, redundancy, or contamination---that are relevant to downstream learning or inference performance. As the foundation of the pipeline, effective data optimization depends on sufficiently accurate perception of the involved data.}

\item \textbf{{Data analysis}:} {This step involves techniques for interpreting the perception results and estimating data defects and {data utility}. In this paper, {data utility} refers to the extent to which a sample, subset, or corpus is useful for improving a target objective of the LLM, such as general performance, domain adaptation, robustness, alignment quality, or inference efficiency. Two primary analysis strategies exist: (1) statistical analysis, which directly computes statistics from the perceived data to identify defects or utilities at the sample, subset, or corpus level; and (2) model-based analysis, which applies statistical modeling to the perception results for training/inference data. In this survey, we do not review data analysis as a standalone section because its methods substantially overlap with the adjacent stages of the pipeline. In existing LLM studies, analysis is often intertwined with data perception when signals are directly interpreted upon measurement, or embedded in data optimization when the resulting scores are immediately used to guide interventions such as selection, reweighting, or augmentation.}

\item \textbf{{Data optimization}:} The data perception and analysis steps serve as preprocessing for the subsequent optimization phase. As the most crucial component of the entire pipeline, this step will be thoroughly examined in Section~\ref{sec5} through our taxonomy of current optimization techniques for LLMs.
\end{itemize}

{Therefore, while data analysis is conceptually indispensable in the proposed pipeline, it is discussed in this survey mainly through its integration with data perception and data optimization, rather than as a fully separate category.}

{\paragraph{Boundary with data synthesis.}
Data optimization is defined as the systematic improvement of \emph{existing} data utilization through operations such as selecting, augmenting, reweighting, re-ordering, or compressing data that are anchored to an already-collected corpus. 
Pure data synthesis---creating training samples from scratch without reference to an existing dataset---falls outside this definition. 
However, when synthetic data is generated conditioned on an existing corpus (e.g., paraphrasing existing instructions, back-translating raw text, or generating counterfactual variants of real samples), it constitutes data augmentation, a core optimization technique. 
Thus, the boundary hinges on whether the operation is anchored to an existing data distribution: optimization transforms or curates existing data, whereas synthesis creates de novo data.}

\section{Data Perception for LLMs}\label{sec4}
Building upon the taxonomy for data perception in conventional deep learning~\cite{wu2024data}, this study examines data perception specifically through the lens of perception types or dimensions. We consider eight key dimensions: quality, difficulty, diversity,  uncertainty, value, redundancy, balance, and others (such as contamination and coverage).

\subsection{Operational Definitions and Dimensional Boundaries}
\label{sec:definitions}
To ensure rigorous application of our taxonomy, we first clarify the 
conceptual boundaries among these dimensions, as they are frequently 
conflated in existing literature. Three critical distinctions warrant 
explicit discussion:
\begin{itemize}
\item {Quality versus Diversity.} These dimensions are \emph{orthogonal}. 
Quality concerns the \emph{correctness} of individual samples (absence of 
noise/errors), while diversity concerns the \emph{variation} across samples 
(coverage of distinct patterns). A dataset can be high-quality but 
low-diversity (e.g., accurate but homogeneous encyclopedic text), or 
high-diversity but low-quality (e.g., broad-topic web text with factual 
errors).
    \item {Difficulty versus Complexity.} Complexity usually serves as a 
\emph{measurable proxy} for difficulty. \emph{Complexity} refers to intrinsic 
sample attributes (reasoning steps, length, syntactic depth), while 
\emph{difficulty} refers to model-dependent learning dynamics (loss, 
perplexity, accuracy). We subsume complexity under difficulty operationally, 
using complexity metrics as difficulty indicators, while acknowledging that 
a sample's difficulty varies with model capability whereas its complexity 
remains fixed.
     \item {Diversity versus Redundancy.} Diversity measures \emph{set-level 
coverage} (breadth of variation), while redundancy measures \emph{pairwise 
similarity} (near-duplication). Critically, high diversity does not imply 
low redundancy: a dataset can cover many topics yet contain duplicates 
within each topic. Conversely, narrow-domain data can be non-redundant if 
each sample is unique.
\end{itemize}

\paragraph{Inter-dimensional dependencies, correlations, and tensions.}
While the eight dimensions are conceptually distinct, they exhibit non-trivial interdependencies in practice. We summarize these relationships along three axes:

\begin{itemize}
    \item \textbf{Synergies.} 
    Quality and diversity jointly promote trustworthiness and fairness: high-quality curation reduces hallucination, while broad diversity mitigates representational bias. 
    Difficulty and value are positively correlated, as appropriately difficult samples often carry high marginal value by eliciting complex reasoning capabilities.

    \item \textbf{Correlations and dependencies.} 
    Uncertainty is partially derived from quality (label noise increases predictive entropy) and from difficulty (harder samples tend to exhibit higher model uncertainty). 
    Redundancy inversely correlates with value: near-duplicate samples typically exhibit low marginal Shapley value because they provide little new information.

    \item \textbf{Tensions and conflicts.} 
    Aggressive quality filtering (e.g., removing all samples with high perplexity) may inadvertently discard rare linguistic patterns or tail-domain knowledge, thereby reducing diversity and fairness. 
    Enforcing strict demographic or domain balance through reweighting can dilute the influence of high-value samples that happen to belong to overrepresented groups, potentially capping peak performance. 
    Over-prioritizing difficult samples can destabilize training dynamics, especially in early stages, because excessively hard examples may contain adversarial noise or ambiguous labels.
\end{itemize}

{Balance and Diversity are related but distinct: a dataset can be diverse in topics yet imbalanced in demographic representation, and vice versa. Their tension is particularly salient in multilingual corpora, where high language diversity may coexist with severe resource imbalance across languages.} These relationships are summarized in Table~\ref{tab:inter-dim}, serving as a compact reference for the interdependencies that practitioners must navigate when composing optimization strategies.

\begin{table}[t]
\caption{Inter-dimensional dependencies, correlations, and tensions.}
\label{tab:inter-dim}
\centering
\small
{
\begin{tabular}{@{}llp{8cm}@{}}
\toprule
\textbf{Relationship type} & \textbf{Dimension pair} & \textbf{Description} \\
\midrule
Synergy & Quality $\times$ Diversity & Jointly promote trustworthiness and fairness \\
Synergy & Difficulty $\times$ Value & Hard samples often carry high marginal value \\
Correlation & Uncertainty $\sim$ Quality & Label noise increases predictive entropy \\
Correlation & Uncertainty $\sim$ Difficulty & Harder samples show higher model uncertainty \\
Correlation & Redundancy $\sim$ Value & Near-duplicates exhibit low Shapley value \\
Tension & Quality vs. Diversity & Aggressive filtering erodes tail-domain coverage \\
Tension & Balance vs. Value & Strict balance dilutes high-value majority samples \\
Tension & Difficulty vs. Training stability & Over-prioritizing hard samples destabilizes training \\
\bottomrule
\end{tabular}}
\end{table}

Table~\ref{tab:perception_dimensions} summarizes operational definitions, 
quantitative proxies, and an intuitive explanation of how each
dimension connects to LLM performance. 
  These eight perception dimensions correspond to the ``data defects'' introduced in Section 1 as follows: quality (low quality), difficulty (insufficiently challenging), diversity (lack of diversity), redundancy (redundancy), balance (imbalance), uncertainty (derived from noise), value (data utility), and others (contamination/coverage). It is important to note that several perception dimensions—particularly quality, difficulty, and value—lack strict, universally accepted definitions. As discussed in Section~\ref{sec2}, diversity and redundancy are frequently employed as indicators of data quality, while uncertainty often serves as a measure of difficulty or value. Rather than attempting to formalize these ambiguous dimensions, {we define each within a specific operational scope while allowing for necessary 
extensions. } 


\begin{table}[t]
\centering
\small
\caption{{Data perception dimensions: definitions, proxies, and relevance to LLM performance.}}
\label{tab:perception_dimensions}\vspace{-0.1in}
{
\begin{tabular}{@{}p{2cm}p{3cm}p{3.5cm}p{5cm}@{}}
\toprule
\textbf{Dimension} & \textbf{Operational Definition} & \textbf{Common Quantitative Proxies} & \textbf{Connection to LLM Performance} \\
\midrule
Quality & The extent to which data is free from noise, errors, and formatting inconsistencies & Perplexity, domain-specific vocabulary ratio, readability scores, annotation agreement & If neglected: model replicates noise, generates factual errors, and degrades instruction-following; if optimized: enhances factual accuracy and output coherence \cite{Gao2024ont,havrilla2024understanding} \\
\midrule
Difficulty & The level of challenge a sample presents to the model's learning or inference process & Loss-based (training/validation loss), perplexity, reasoning steps length, reference model accuracy & If too low: model resorts to simple pattern matching, failing on complex tasks; if too high: training becomes unstable and convergence difficult; if optimized: elicits complex reasoning and prevents premature convergence \cite{chen2024bull,gao2025prin,muennighoff2025s1} \\
\midrule
Diversity & The degree of variation and coverage across samples in a dataset & Vendi score (entropy of similarity matrix eigenvalues), cluster counts, pairwise distances, lexical diversity & If insufficient: fails on minor domains/styles/groups, overfits to training patterns; if optimized: improves cross-domain generalization and reduces spurious correlations \cite{havrilla2024survey,li2025eagle,lu2023instag} \\
\midrule
Uncertainty & The level of ambiguity in model predictions or noise in training labels & Predictive entropy, semantic consistency across samples, LLM self-reported confidence & If unquantified: overconfident in wrong answers, unaware of knowledge boundaries; if leveraged: enables abstention or human review, mitigating hallucinations \cite{Farquhar2024,kadavath2022lan,shorinwa2024survey} \\
\midrule
Value & The marginal contribution of a sample to model performance & Shapley value approximations, influence functions, gradient-based influence, training-free proxies & If neglected: training inefficient, compute wasted on uninformative samples; if optimized: achieves comparable performance with less data, enables precise demonstration selection \cite{kwon2024data,sachdeva2024train,xia2024less} \\
\midrule
Redundancy & The presence of highly similar or duplicated content across samples & N-gram overlap, embedding similarity (e.g., within epsilon balls), information gain scores & If unaddressed: model memorizes rather than learns generalizable patterns; if optimized: improves both efficiency and generalization \cite{Cunha2025,Tirumala2023,yin2024entropy} \\
\midrule
Balance & The equitable representation of different groups, topics, languages, or attributes & Class distribution statistics, demographic parity metrics, language representation ratios & If imbalanced: performance disparities across minority groups/languages/topics, fairness issues; if optimized: balanced performance across subgroups, improved robustness \cite{huang2025survey,jones2025exam,ju2024mit} \\
\midrule
Others (Contam./Coverage) & Unintended train--test overlap or insufficient coverage of deployment-relevant domains, tasks, or inputs & Overlap rate, duplicate match statistics, embedding similarity, domain/task coverage ratios & If neglected: benchmark scores inflated or robustness/generalization weakened; if optimized: more trustworthy evaluation and broader deployment reliability \cite{balloccu2024,dong2024gen,singh2024eva} \\
\bottomrule
\end{tabular}}
\end{table}

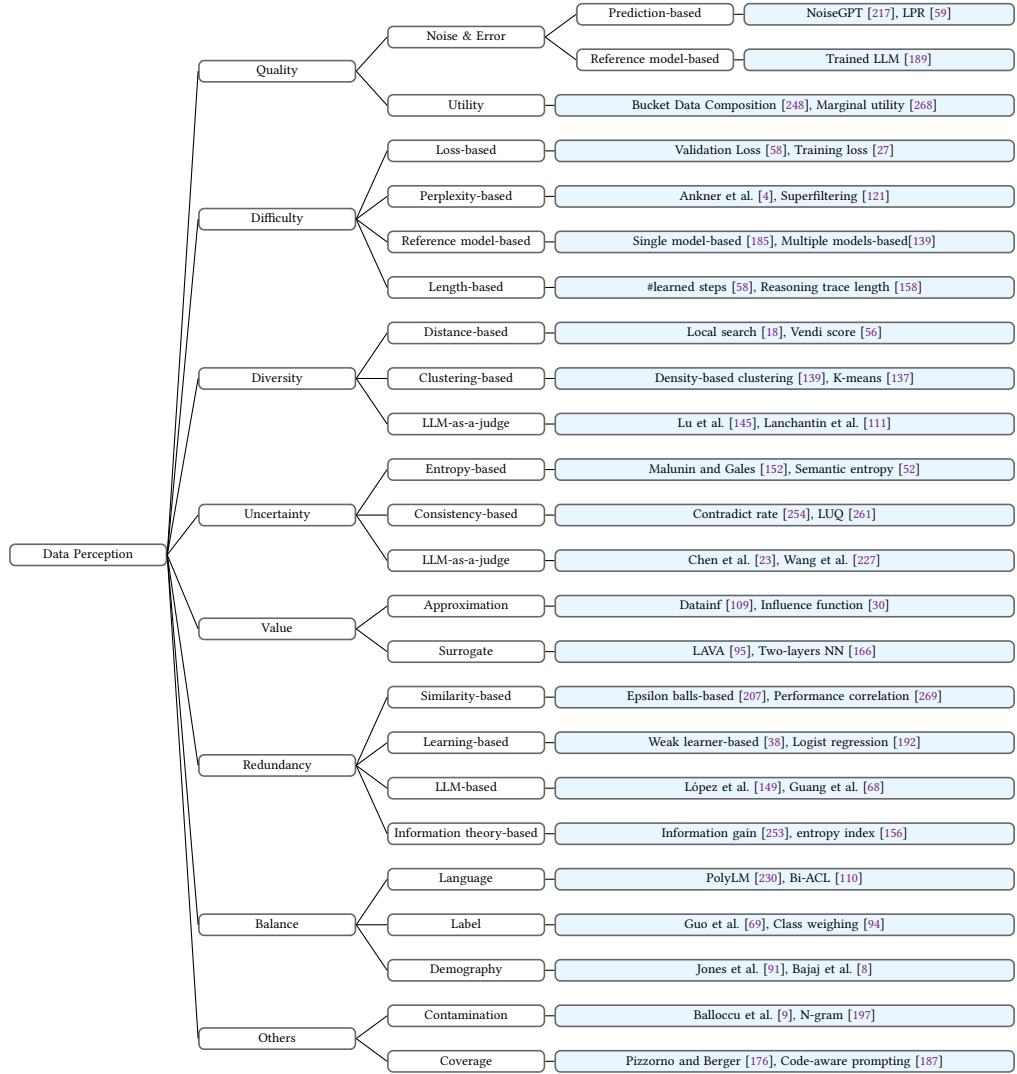
\begin{figure*}[t]
\begin{center}
\begin{tikzpicture}
 
\tikzset{
  grow'=right,level distance=25mm, sibling distance =3mm,
  execute at begin node=\strut,
  every tree node/.style={
			draw=gray!80!black,
            line width=0.6pt,
            text width=2cm,
			rounded corners=2pt,
			anchor = west,
            fill=white,
			minimum width=5mm,
			inner sep=1pt,
			align=center,
			font = {\tiny}}
}
\begin{scope}[frontier/.style={sibling distance=4em,level distance = 7em}]
\Tree
[.{Data Perception}
		[.{Quality}
			[.{Noise \& Error}
               [.{Prediction-based}
              \node[fill=CY,text width=3.5cm](t1){NoiseGPT~\cite{wang2024noisegpt}, LPR~\cite{Gao2024ont}};
              ]
               [.{Reference model-based}
              \node[fill=CY,text width=3.5cm](t1){Trained LLM~\cite{sachdeva2024train}};
              ]
			]
			[.{Utility}
              \node[fill=CY,text width=6cm](t1){Bucket Data Composition~\cite{liu2024bucket}, Marginal
 utility~\cite{zhang2022act}};
			]
		]
		[.{Difficulty}
              [.{Loss-based}
              \node[fill=CY,text width=6cm](t1){Validation Loss~\cite{gao2025prin}, Training loss~\cite{chen2024bull}};
              ]
              [.{Perplexity-based}
              \node[fill=CY,text width=6cm](t1){Ankner et al.~\cite{ankner2024per}, Superfiltering~\cite{li2024super}};
              ]
              [.{Reference model-based}
              \node[fill=CY,text width=6cm](t1){Single model-based~\cite{rogoz2024unib}, Multiple models-based\cite{liu-etal-2025-empirical}};
              ]
              [.{Length-based}
              \node[fill=CY,text width=6cm](t1){\#learned steps~\cite{gao2025prin}, Reasoning trace length~\cite{muennighoff2025s1}};
              ]
		]
       [.{Diversity}
              [.{Distance-based}
              \node[fill=CY,text width=6cm](t1){Local search~\cite{cevallos2016local}, Vendi score~\cite{friedman2023vend}};
              ]
              [.{Clustering-based}
              \node[fill=CY,text width=6cm](t1){Density-based clustering~\cite{liu-etal-2025-empirical}, K-means~\cite{ling2025div}};
              ]
              [.{LLM-as-a-judge}
              \node[fill=CY,text width=6cm](t1){Lu et al.~\cite{lu2023instag}, Lanchantin et al.~\cite{lanchantin2025diver}};
              ]
       ]
       [.{Uncertainty}
              [.{Entropy-based}
              \node[fill=CY,text width=6cm](t1){Malunin and Gales~\cite{malinin2021uncer}, Semantic entropy~\cite{Farquhar2024}};
              ]
              [.{Consistency-based}
              \node[fill=CY,text width=6cm](t1){Contradict rate~\cite{yona2024large}, LUQ~\cite{zhang2024luq}};
              ]
              [.{LLM-as-a-judge}
              \node[fill=CY,text width=6cm](t1){Chen et al.~\cite{chen2023quan}, Wang et al.~\cite{wang2024uncert}};
              ]
       ]
       [.{Value}
              [.{Approximation}
              \node[fill=CY,text width=6cm](t1){Datainf~\cite{kwon2024data}, Influence function~\cite{choe2024data}};
              ]
              [.{Surrogate}
              \node[fill=CY,text width=6cm](t1){LAVA~\cite{just2023lava}, Two-layers NN~\cite{nohyun2023data}};
              ]
       ]
       [.{Redundancy}
	         [.{Similarity-based}
              \node[fill=CY,text width=6cm](t1){Epsilon balls-based~\cite{Tirumala2023}, Performance correlation~\cite{zhang2025red}};
              ]
              [.{Learning-based}
              \node[fill=CY,text width=6cm](t1){Weak learner-based~\cite{Cunha2025}, Logist regression~\cite{Shah2024how}};
              ]
              [.{LLM-based}
              \node[fill=CY,text width=6cm](t1){López et al.~\cite{López2025}, Guang et al.~\cite{10808217}};
              ]
              [.{Information theory-based}
              \node[fill=CY,text width=6cm](t1){Information gain~\cite{yin2024entropy}, entropy index~\cite{mitchell2023meas}};
              ]
       ]
       [.{Balance}
	         [.{Language}
              \node[fill=CY,text width=6cm](t1){PolyLM~\cite{wei2023poly}, Bi-ACL~\cite{lai2023mit}};
              ]
              [.{Label}
              \node[fill=CY,text width=6cm](t1){Guo et al.~\cite{guo2024makes}, Class weighing~\cite{jung2024und}};
              ]
              [.{Demography}
              \node[fill=CY,text width=6cm](t1){Jones et al.~\cite{jones2025exam}, Bajaj et al.~\cite{bajaj2024eva}};
              ]
       ]
       [.{Others}
	         [.{Contamination}
              \node[fill=CY,text width=6cm](t1){Balloccu et al.~\cite{balloccu2024}, N-gram~\cite{singh2024eva}};
              ]
              [.{Coverage}
              \node[fill=CY,text width=6cm](t1){Pizzorno and Berger~\cite{pizzorno2025cov}, Code-aware prompting~\cite{Ryan2024Cod}};
              ]
       ]
]
\end{scope}
\end{tikzpicture}
\caption{A taxonomy for data perception for LLMs. {\textbf{Reading guide:} The tree is read from root (left) to leaves (right). Each branch under ``Data Perception'' corresponds to one of the eight dimensions discussed in Section 4. Second-level nodes indicate measurement families (e.g., ``Loss-based'' under Difficulty), and leaf nodes cite representative methods. The taxonomy is operational: it maps abstract dimensions to concrete, computable proxies.}}
\label{fig3}
\end{center}\vspace{-0.1in}
\end{figure*}

Fig.~\ref{fig3} presents an overview of the data perception framework examined in this study. In the following subsections, we elaborate on each dimension, reviewing representative measurement methods and highlighting their roles in the LLM lifecycle.

\subsection{Data Quality}
Following the study in ~\cite{havrilla2024survey}, this dimension aims to assess potential noise and errors spanning from individual training samples to the entire dataset, with primary consideration given to both the samples themselves and their corresponding target values~(particularly in standard supervised learning contexts). For instance, the textual data for many LLMs are typically Web-sourced and may contain substantial inaccuracies and non-standard expressions. In regression tasks where target values represent continuous scores, these labels are susceptible to measurement noise. Conversely, for classification tasks employing discrete categorical labels, the primary concern resides in potential mislabeling errors within class assignments.

Numerous studies exist on measuring noise and errors in training samples. Typical methods include prediction-based and reference model-based approaches. Wang et al.~\cite{wang2024noisegpt} generated a set of perturbations for each sample and leveraged the predictions of all perturbed samples to identify noisy labels. Sachdeva et al.~\cite{sachdeva2024train} applied a trained LLM to assess the quality of individual training samples. For example, Nahum et al.~\cite{nahum2024llms} evaluated the technique of LLM-as-a-judge to delete potentially mislabeled examples. Havrilla and Iyer~\cite{havrilla2024understanding} revealed that low or moderate amounts of noise may result in robust LLM performance in chain-of-thought reasoning. Gan et al.~\cite{gan2024ziya2} proposed a framework for automated quality assessment of large-scale datasets, introducing quantifiable metrics including domain-specific vocabulary ratio, information entropy, readability metrics, linguistic distribution patterns, and document length statistics, accompanied by concrete quantification methodologies for each dimension. For example, domain-specific vocabulary ratio is calculated as the number of domain keywords (e.g., medical terms for biomedical data) divided by the total number of tokens in the dataset. Gao et al.~\cite{Gao2024ont} demonstrated that noisy annotations significantly impair the performance of in-context learning on text generation tasks, leveraging perplexity to identify noisy data.

Extensions to conventional data quality measures primarily focus on the utility of training data. For instance, Yang et al.~\cite{liu2024bucket} established a novel quality measure for training corpus composition consisting of three indicators: padding ratio, truncation ratio, and concatenation ratio. The first two indicators reflect the information density of training data, while the third reflects inconsistency introduced by manual concatenation. Zhang et al.~\cite{zhang2022act} measured the marginal utility of candidate examples in in-context learning.

\paragraph{Connection to LLM Performance}
{Data quality exerts a direct and profound influence on LLM behavior. Low-quality data—such as factual inaccuracies, grammatical errors, or formatting inconsistencies—propagates to model outputs, manifesting as factual hallucinations, reduced coherence, and degraded instruction-following ability \cite{Gao2024ont,havrilla2024understanding}. Conversely, high-quality data curation, even at smaller scales, can unlock capabilities that rival those trained on much larger but noisier corpora \cite{abdin2024phi4}. The impact is particularly pronounced in specialized domains: for instance, noisy annotations in medical or legal text can lead to clinically significant errors, underscoring the non-negotiable nature of quality in high-stakes applications. Quality perception thus serves as the first line of defense against ``garbage in, garbage out''.}

\subsection{Data Difficulty}
This dimension aims to assess the learning/inference difficulty of individual samples. Data difficulty, quality, and diversity constitute the three primary dimensions in LLM data perception. As with conventional deep learning, two fundamental learning approaches—curriculum learning and self-paced learning—entirely depend on data difficulty measurement. Effective difficulty assessment is crucial for evaluating LLM performance across tasks and profiling its easy-to-hard generalization capabilities~\cite{ding2024easyhardbench}. No universally accepted rigorous definition of data difficulty currently exists. Zhu et al.~\cite{ZhuTKDD} proposed a formal definition and approximate measurement method based on generalization error analysis, though its computational complexity remains prohibitive for LLM applications.

In conventional deep learning, five primary methods measure sample difficulty: loss-based, gradient-based, uncertainty-based, multi-training-based, and distance-based approaches. The multi-training-based approach proves impractical for LLMs. While conventional deep learning primarily focuses on classification, LLMs serve general purposes, making difficulty measurements context-dependent. For LLMs, four main measurement types exist: loss-based, perplexity-based, reference model-based, and length-based approaches. Gao et al.~\cite{gao2025prin} utilized validation loss as a difficulty signal and adopted an ``increasing difficulty'' selection strategy (prioritizing samples with higher validation loss). This avoids premature convergence during LLM alignment and improves generalization on complex tasks. Chen et al.~\cite{chen2024bull} employed training loss as a difficulty metric during pretraining. Perplexity\footnote{We note that perplexity should be interpreted as a proxy for difficulty rather than a strict definition of it. Although useful and scalable, it may also reflect domain shift or model-specific effects, and thus may not fully capture reasoning difficulty on its own.} serves as a widely-used difficulty metric for tasks like pretraining data selection~\cite{ankner2024per} and instruction tuning~\cite{li2024super}. Reference model-based measurements may employ single or multiple models. Rogoz and Ionescu~\cite{rogoz2024unib} trained a model to predict sample difficulty using heuristic features for Multiple-Choice Question tasks. Liu et al.~\cite{liu-etal-2025-empirical} employed twenty models to measure questions’ difficulties through accuracy distributions. Ye et al.~\cite{ye2025li} similarly employed LLMs to assess sample difficulty.

Alternative measurement schemes include using the number of learning steps to measure difficulty in LLM alignment~\cite{gao2025prin}. The related dimension of data complexity also appears in LLM studies. Liu et al.~\cite{liu2024what} examined instruction tuning data selection using complexity, quality, and diversity dimensions, where complexity scoring incorporated perplexity, instruction length, response loss, and reasoning steps—factors commonly associated with difficulty measurement. Havrilla et al.~\cite{havrilla2024survey} noted that complexity measures intuitively reflect ``difficulty'' or ``compositionality.'' Since complexity~(e.g., reasoning steps) is a key sub-dimension of learning difficulty for LLMs, we merge complexity into the difficulty dimension. Li et al.~\cite{li2024quant} defined instruction-following difficulty as a novel metric, while Muennighoff et al.~\cite{muennighoff2025s1} used reasoning trace length as a difficulty indicator for mathematical problem-solving tasks.

\paragraph{Connection to LLM Performance}
{The difficulty distribution of training data fundamentally shapes the reasoning capabilities of LLMs. Exposure to insufficiently challenging samples---such as short, template-based instructions---can lead to premature convergence and weak generalization on complex tasks \cite{gao2025prin}. Conversely, appropriately difficult samples, particularly those requiring multi-step reasoning or chain-of-thought generation, are essential for eliciting and reinforcing systematic reasoning capabilities \cite{muennighoff2025s1}. Recent work on ``easy-to-hard'' generalization demonstrates that models trained predominantly on easy examples struggle to generalize to harder distributions, while strategic inclusion of difficult examples at key training stages can substantially improve performance on challenging benchmarks \cite{ding2024easyhardbench}. Difficulty-aware data selection thus functions as a lever for targeted capability enhancement.}

\subsection{Data Diversity}
As defined in \cite{havrilla2024survey}, data diversity intuitively measures the ``self-similarity'' or ``coverage'' of data. This dimension ranks among the most widely used factors in previous studies. Some researchers even prioritize diversity over quality, considering it the foremost criterion~\cite{li2025eagle,xia2024rethink}. Unlike quality and difficulty, diversity primarily applies to sample sets rather than individual samples. Three main schemes exist for measuring data diversity: distance-based, clustering-based, and LLM-as-a-judge approaches.

The distance-based scheme calculates pairwise similarity or distance between samples. Cevallos et al.~\cite{cevallos2016local} defined dispersion notation as the sum of all pairwise distances to quantify dataset diversity. Friedman and Dieng~\cite{friedman2023vend} designed the Vendi score metric, defined as the exponential of the Shannon entropy of similarity matrix eigenvalues.

Clustering-based schemes represent an enhanced version of distance-based methods. This approach typically involves two steps: first clustering samples into groups, then calculating statistical quantities for each group to assess diversity. Liu et al.~\cite{liu-etal-2025-empirical} employed a density-based clustering algorithm to partition data, using question/topic counts and their variance/standard deviation as diversity measures.

The LLM-as-a-judge approach utilizes one or more LLMs to evaluate dataset diversity. Lu et al.~\cite{lu2023instag} employed ChatGPT to generate detailed tags measuring instruction variety during SFT. Lanchantin et al.~\cite{lanchantin2025diver} prompted an LLM to identify the most and least diverse responses within given data.

Additional diversity measures include task-oriented metrics summarized by Mitchell et al.~\cite{mitchell2023meas}
~(lexical diversity, inception diversity) and Yu et al.'s~\cite{yu-etal-2022-data} convex hull volume approach for measuring sentence set coverage.

\paragraph{Connection to LLM Performance}
{Data diversity is a key determinant of model generalization and robustness. Low-diversity training corpora, dominated by a few linguistic patterns or topics, cause models to learn spurious correlations and underperform on out-of-distribution inputs \cite{havrilla2024survey}. In contrast, diverse data exposes the model to a wide range of linguistic variations, domains, and tasks, improving its ability to adapt to new contexts and reducing reliance on superficial cues \cite{lu2023instag,li2025eagle}. For multilingual and multimodal LLMs, diversity across languages and modalities is particularly critical to avoid biased representations, promote fairness, and ensure equitable performance. Diversity-aware data selection and augmentation are therefore essential for building generally capable and fair models.}

\subsection{Data Uncertainty}
Uncertainty constitutes a fundamental machine learning concept that underpins methodological frameworks~\cite{Eyke2021}. In machine learning, data uncertainty typically refers to training sample noise/error or model prediction confidence. For LLMs, training sample uncertainty follows conventional deep learning paradigms, while most studies focus on output confidence~\cite{shorinwa2024survey}. Uncertainty measurement proves crucial for LLM factuality analysis~\cite{10820047}. Shorinwa et al.~\cite{shorinwa2024survey} categorized uncertainty quantification into token-level, self-verbalized, and semantic-similarity methods. We propose an alternative taxonomy: entropy-based, consistency-based, and LLM-as-a-judge approaches.

Entropy-based methods derive from conventional deep learning. Malinin and Gales~\cite{malinin2021uncer} investigated information-theoretic uncertainty measures based on entropy, while Farquhar et al.~\cite{Farquhar2024} developed semantic entropy by clustering answers semantically before entropy calculation.

Consistency-based methods compute response variation for given queries, using inconsistency scores as uncertainty measures~\cite{yona2024large}. Zhang et al.~\cite{zhang2024luq} evaluated long-response quality through pairwise response similarities (consistency).

The LLM-as-a-judge approach sees widespread practical use. Chen et al.~\cite{chen2023quan} introduced self-reflection certainty, querying LLMs about their response confidence. Wang et al.~\cite{wang2024uncert} employed superior models (e.g., GPT-4) for uncertainty modeling.

Other approaches include Liu et al.'s~\cite{liu2024uncer} trained uncertainty prediction model and investigations into hidden state information. Kadavath et al.~\cite{kadavath2022lan} predicted response confidence from hidden states, while Plaut et al.~\cite{plaut2025prob} demonstrated maximum softmax probabilities as effective uncertainty measures.

\paragraph{Connection to LLM Performance}
{Uncertainty quantification plays a vital role in enhancing the reliability and trustworthiness of LLMs. By identifying inputs where the model's predictions are uncertain, practitioners can flag potential hallucinations, abstain from answering when confidence is low, or trigger human review \cite{Farquhar2024,shorinwa2024survey}. Uncertainty estimates also enable selective generation, where models are trained to recognize the limits of their knowledge, thereby reducing overconfidence and improving factual accuracy \cite{kadavath2022lan}. Furthermore, uncertainty-aware sampling can guide active learning and data annotation efforts, focusing human effort on examples where the model is most uncertain and thus most likely to benefit from additional training.}

\subsection{Data Value}
Data valuation for LLMs aims to identify the most valuable or influential samples or demonstrations in a learning or an inference task. In conventional machine learning, data value has a formal definition based on the Shapley value from game theory~\cite{pmlrv97}. Nevertheless, the calculation of the Shapley value is NP-hard, thereby hindering its application in real scenarios. In the era of LLMs, data valuation methods can be roughly divided into two folds, namely, efficient approximations and training-free surrogates~\cite{xu2024data}.

To reduce the computational complexity of the Shapley value, a value transfer method is developed that leverages value information extracted from a simple classifier trained using representations from the target language model~\cite{schoch2023data}. Kwon et al.~\cite{kwon2024data} proposed an efficient data influence approximation method that has theoretical guarantees. Choe et al.~\cite{choe2024data} designed an efficient gradient projection strategy to accelerate the popular gradient-based data valuation strategy. Xia et al.~\cite{xia2024less} also investigated a practically efficient algorithm to effectively estimate data influences for instruction tuning. Kong et al.~\cite{Kong2022Res} evaluated the harmfulness of each training sample based on influence functions. Wang et al.~\cite{wang2025data} proposed a quite efficient Shapley-based valuation algorithm based on a validation dataset.

Training-free surrogates employ a surrogate function to implement scalable valuation. Just et al.~\cite{just2023lava} developed a proxy for the validation performance associated with a training set based on Wasserstein distance. Nohyun et al.~\cite{nohyun2023data} provided a training-free valuation score to quantify the influence of individual instances in training. Wu et al.~\cite{pmlrwu22j} derived a domain-aware theory to estimate the generalization performance of DNNs without training and applied it to develop a novel training-free data valuation method. 

There are some studies focusing on different scenarios or strategies. Xie et al.~\cite{xie2024demo} investigated the assessment of the influence of individual demonstrations in in-context learning based on the Shapley value, as the computational cost is relatively low in this scenario. Jiao et al.~\cite{jiao2025feasi} explored the feasibility of prompting an LLM to serve as a fast proxy for gradient-based data valuation.

\paragraph{Connection to LLM Performance}
{Data valuation identifies training (or inference) examples that contribute most to model performance, enabling efficient data pruning and targeted acquisition. Removing low-value samples reduces computational costs without sacrificing accuracy, while prioritizing high-value samples accelerates convergence and improves final performance \cite{kwon2024data,xia2024less}. In instruction tuning and few-shot learning, value estimation helps select the most influential demonstrations, maximizing the impact of limited data \cite{xie2024demo}. Moreover, value-aware data reweighting can mitigate the influence of harmful or outlier samples, enhancing model robustness and alignment. Data valuation thus underpins both efficiency and effectiveness in LLM development.}

\subsection{Data Redundancy}
Compared with the above-mentioned perception dimensions, this dimension receives less attention. Nevertheless, it is also quite crucial for the training of LLMs. Redundancy for LLMs refers to the presence of highly similar or duplicate questions that may lead to a waste of computational resources~\cite{liu-etal-2025-empirical}. Indeed, redundancy can also degrade model performance and impact the generalization ability and prediction accuracy of LLMs~\cite{10808217}.

Chen et al.~\cite{10.1145/3564663} published a comprehensive survey on data redundancy for deep learning. They summarize four ways for redundancy detection, namely, similarity-based, learning-based, projection-based, and implicit detection-based. The similarity-based strategy is also used for LLMs. For example, Tirumala et al.~\cite{Tirumala2023} detected duplicated samples if they are within epsilon balls of one another. Cunha et al.~\cite{Cunha2025} estimated data redundancy based on scalable, fast, and calibrated weak classifiers (e.g., logistic regression). The projection-based and implicit detection-based strategies are rarely used for LLMs due to computational cost considerations.

There are new explorations particularly for LLMs. Guang et al.~\cite{10808217} investigated how to eliminate redundancy for training samples based on LLM prompt engineering and sequential optimization. Methods based on information theory are also investigated. Yin et al.~\cite{yin2024entropy} maintained an information redundancy state that evaluates the ``information gain'' of each sample. A subset of samples with the lowest scores is selected to compress the raw training corpus. Mitchell et al.~\cite{mitchell2023meas} stated that entropy can also be used to measure redundancy, such as the generalized entropy index.

\paragraph{Connection to LLM Performance}
{Data redundancy, in the form of near-duplicate or highly similar examples, degrades LLM training in multiple ways. It wastes computational resources by forcing the model to process repeated information, slowing convergence and increasing training costs \cite{Tirumala2023}. More critically, redundancy can cause models to over-memorize rather than generalize, leading to reduced performance on unseen data \cite{yin2024entropy}. Deduplication and redundancy-aware pruning have been shown to improve both efficiency and generalization, sometimes even boosting accuracy by encouraging the model to focus on diverse patterns \cite{Cunha2025}. Detecting and eliminating redundancy is therefore a fundamental step in data optimization for LLMs.}

\subsection{Data Balance}
The issue of data imbalance is prevalent in both the training and inference phases. Pretraining relies on massive training datasets aggregated from diverse sources that vary significantly in style, content, and other attributes. Each individual source typically encompasses heterogeneous topics, languages, and linguistic characteristics, all of which collectively contribute to systemic imbalance. This inherent heterogeneity in data composition across sources manifests as distributional disparities in feature representation, thereby introducing bias amplification risks during model optimization and subsequent inference tasks. In the context of LLMs, the following types of imbalance receive more attention than others, namely language imbalance, label imbalance, and demographic imbalance.

Many studies attempt to deal with the imbalance issue among resources with different languages in LLM training~\cite{huang2025survey}. PolyLM~\cite{wei2023poly} integrates bilingual alignment for training data and employs curriculum learning to maintain a balance among various languages during pretraining. Guo et al.~\cite{guo2024makes}  proposed an adaptive strategy for few-shot in-context learning: first filtering orderings that cause prediction imbalance, then selecting the ordering with maximum impact on test samples. This goes beyond generic class-balance filtering to optimize in-context example sequencing. Jones et al.~\cite{jones2025exam} analyzed demographic imbalances in a standard benchmark disease dataset across gender, age, ethnicity, and social determinants of health by SOTA biomedical language models. They revealed that data imbalance significantly impacts model performance and fairness.

The training order or positions of training samples heavily influence the final model performance. Ju et al.~\cite{ju2024mit} explicitly revealed order imbalance in SFT: samples processed at different positions unevenly influence the learning process, and this imbalance poses a potential risk of skewing the performance of the fine-tuned model.

As a main factor resulting in bias is imbalance among demographic attributes, bias measurement is also placed into this part. Lin et al.~\cite{Lin2024Tow} presented a systematic survey for trustworthy LLMs, including the bias issue. A set of bias evaluation metrics are discussed and presented. Chu et al.~\cite{10.1145/3682112.3682117} summarized three bias metrics, namely, embedding-based, probability-based, and generation-based.

\paragraph{Connection to LLM Performance}
{Imbalance in training data directly translates to skewed model behavior and fairness violations. Language imbalance causes models to perform poorly on low-resource languages \cite{huang2025survey}; demographic imbalance leads to biased predictions across gender, race, or age groups \cite{jones2025exam}; and task imbalance results in over-specialization on frequent tasks while neglecting rare but important ones \cite{ju2024mit}. Addressing imbalance through resampling, reweighting, or synthetic data generation not only improves fairness but also enhances overall robustness by preventing the model from overfitting to majority patterns. Balanced data is a prerequisite for equitable and generalizable LLMs.}

\subsection{Other Dimensions}
There are studies focusing on other perception dimensions, such as contamination and coverage. These dimensions are highly correlated to the seven ones described above.

Balloccu et al.~\cite{balloccu2024} conducted the first systematic analysis of work using OpenAI's GPT-3.5 and GPT-4 and revealed that these models have been globally exposed to a large number of contaminated samples. Singh et al.~\cite{singh2024eva} proposed novel n-gram based contamination metrics. Dong et al.~\cite{dong2024gen} pointed out significant challenges in detecting and mitigating data contamination and presented a new contamination detection method via output distribution.

Data coverage is mainly investigated for code LLMs. Pizzorno and Berger~\cite{pizzorno2025cov} employed coverage measurement to facilitate generating more effective test cases for software development. Tufano et al.~\cite{tufano2023pre} emphasized that code coverage is valuable for overall LLM performance and proposed a novel benchmark task called code coverage prediction for LLMs. Ryan et al.~\cite{Ryan2024Cod} constructed a code-aware prompting strategy for LLMs in test generation with the guidance of coverage. Gupta et al.~\cite{gupta2023cov} defined two coverage measures for a single example and a set, respectively.

\paragraph{Connection to LLM Performance}
{Data contamination and coverage, though distinct, jointly influence the reliability and generalizability of LLMs. Contamination---the unintended inclusion of evaluation data in the training corpus---inflates benchmark scores and masks true model capabilities, undermining trustworthy evaluation \cite{balloccu2024,dong2024gen,singh2024eva}. Coverage, conversely, determines whether training data sufficiently spans the input space encountered at deployment; poor coverage leads to brittle models that fail on out-of-distribution scenarios, particularly in specialized domains like code generation or medicine \cite{pizzorno2025cov,tufano2023pre,Ryan2024Cod}. Together, contamination threatens the validity of performance estimates while inadequate coverage limits real-world utility, making both dimensions essential for responsible LLM development.}



\begin{figure*}[t]
    \centering
    \includegraphics[width=0.98\textwidth]{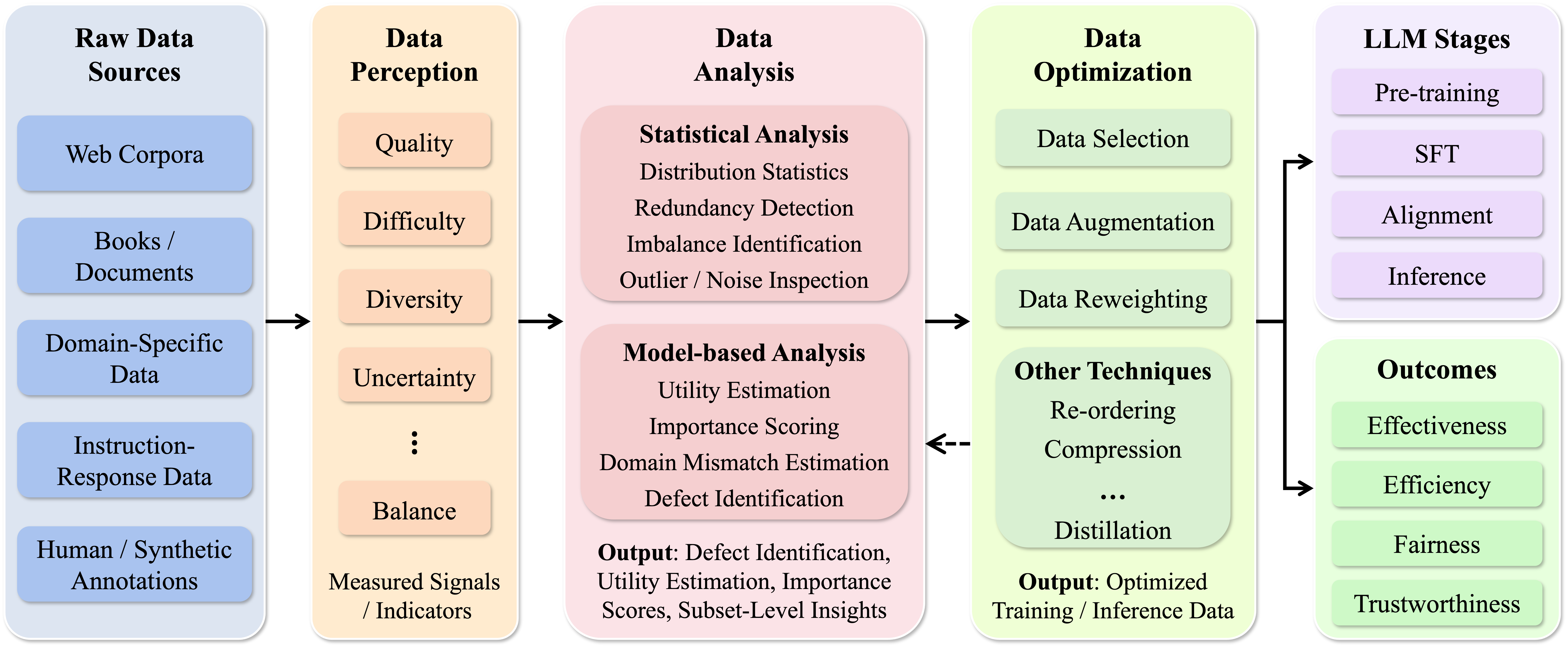}\vspace{-0.1in}
\caption{{An illustrative end-to-end pipeline of data optimization for LLMs.}}

    \label{fig4-0}
\end{figure*}

\section{Data Optimization Techniques}\label{sec5}
{Most optimization methods reviewed in this section rely on an implicit or explicit analysis stage, in which perceived data signals are converted into actionable scores, rankings, or utility estimates.}

\paragraph{Illustrative end-to-end pipeline.}
{Consider a domain-adaptive LLM development scenario. A practitioner may first collect a large mixed corpus from general web data, domain-specific documents, and instruction-response pairs. In the \textit{data perception} step, the corpus is measured along dimensions discussed previously and
other specific dimensions such as compliance. In the \textit{data analysis} step, these signals are converted into actionable estimates, for example by ranking samples according to informativeness, identifying redundant or low-quality instances, and locating underrepresented domains or skills. The \textit{data optimization} step then applies targeted interventions, such as selecting high-value subsets for training, augmenting scarce domain examples, reweighting important samples, or filtering unsafe content. The optimized data are subsequently used in pre-training, SFT, or alignment, and the resulting model is evaluated in terms of effectiveness, efficiency, and trustworthiness. Such a workflow highlights how data optimization serves as a bridge from raw data characteristics to downstream LLM performance, as illustrated in Fig.~\ref{fig4-0}.
}

\subsection{{Overview and Lifecycle Perspective}}
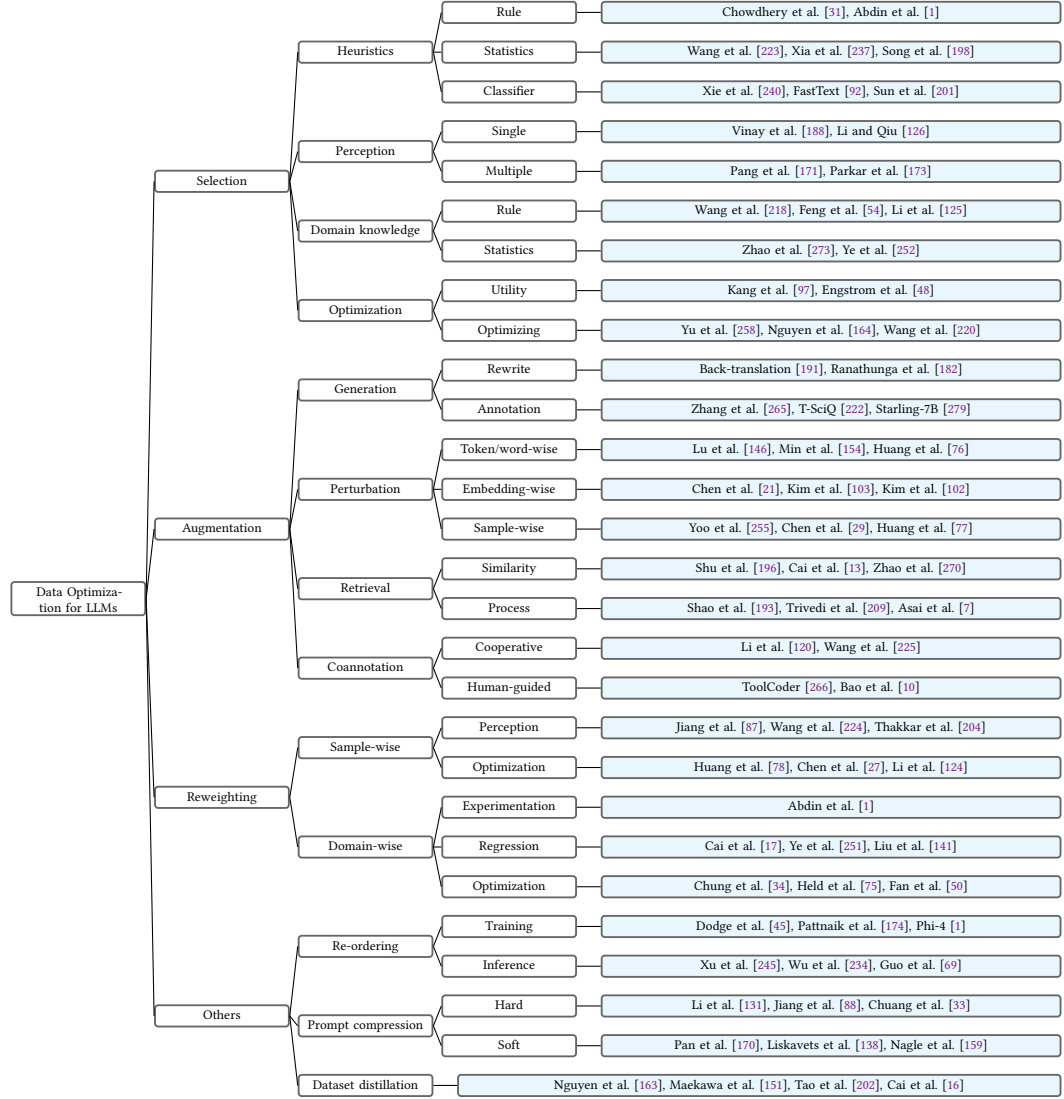
\begin{figure*}[t]
\begin{center}
\scriptsize
\begin{tikzpicture}
 
\tikzset{
  grow'=right,level distance=19mm, sibling distance =2.2mm,
  execute at begin node=\strut,
  every tree node/.style={
			draw=gray!80!black,
            line width=0.7pt,
            text width=1.7cm,
			rounded corners=1.2pt,
			anchor = west,
            fill=white,
			minimum width=0.1mm,
			inner sep=1pt,
			align=center,
			font = {\tiny}}
}
\begin{scope}[frontier/.style={sibling distance=1em,level distance = 10em}]
\Tree
[.{Data Optimization for LLMs}
		[.{Selection}
            [.{Heuristics}
			    [.{Rule}
                  \node[fill=CY,text width=6cm](t1){Chowdhery et al.~\cite{Chowdhery2023}, Abdin et al.~\cite{abdin2024phi4}};
			    ]
			    [.{Statistics}
                  \node[fill=CY,text width=6cm](t1){Wang et al.~\cite{wang2025lan}, Xia et al.~\cite{xia2024rethink}, Song et al.~\cite{song2024lar}
};
			    ]
   		    [.{Classifier}
                  \node[fill=CY,text width=6cm](t1){Xie et al.~\cite{Xie2023Adv}, FastText~\cite{joulin2016fast}, Sun et al.~\cite{sun2024hun}};
			    ]
            ]
            [.{Perception}
			    [.{Single}
                  \node[fill=CY,text width=6cm](t1){Vinay et al.~\cite{s2024inc}, Li and Qiu~\cite{Li2023Find}};
			    ]
			    [.{Multiple}
                  \node[fill=CY,text width=6cm](t1){Pang et al.~\cite{pang2025imp}, Parkar et al.~\cite{parkar2024sel} };
			    ]
            ]
            [.{Domain knowledge}
			    [.{Rule}
                  \node[fill=CY,text width=6cm](t1){Wang et al.~\cite{wang2024surveyd}, Feng et al.~\cite{feng2022aut}, Li et al.~\cite{li2024rul}};
			    ]
			    [.{Statistics}
                  \node[fill=CY,text width=6cm](t1){Zhao et al.~\cite{Zhao2024clue}, Ye et al.~\cite{ye2025li}};
			    ]
            ]
            [.{Optimization}
			    [.{Utility}
                  \node[fill=CY,text width=6cm](t1){Kang et al.~\cite{kang2024less}, Engstrom et al.~\cite{engstrom2024ds}};
			    ]
       		[.{Optimizing}
                  \node[fill=CY,text width=6cm](t1){Yu et al.~\cite{yu2025data}, Nguyen et al.~\cite{nguyen2025mini}, Wang et al.~\cite{Wang2024great}};
			    ]
            ]
		]
		[.{Augmentation}
              [.{Generation}
                  [.{Rewrite}
                    \node[fill=CY,text width=6cm](t1){Back-translation~\cite{Sennrich2016}, Ranathunga et al.~\cite{Ranathunga2023NM}};
                  ]
                  [.{Annotation}
                    \node[fill=CY,text width=6cm](t1){Zhang et al.~\cite{Zhang2024Ult}, T-SciQ~\cite{Wang2024Tsc}, Starling-7B~\cite{zhu2024star}};
                  ]
              ]
              [.{Perturbation}
                  [.{Token/word-wise}
                    \node[fill=CY,text width=6cm](t1){Lu et al.~\cite{Lu2022Ins}, Min et al.~\cite{min2020syn}, Huang et al.~\cite{Hu2024Pro}
};
                  ]
                  [.{Embedding-wise}
                    \node[fill=CY,text width=6cm](t1){Chen et al.~\cite{chen2022dou}, Kim et al.~\cite{kim2023dis}, Kim et al.~\cite{kim2024ins}
};
                  ]
                  [.{Sample-wise}
                    \node[fill=CY,text width=6cm](t1){Yoo et al.~\cite{yoo2021gpt}, Chen et al.~\cite{chen2023dis}, Huang et al.~\cite{huang2024enh}
};
                  ]
              ]
              [.{Retrieval}
                  [.{Similarity}
                    \node[fill=CY,text width=6cm](t1){Shu et al.~\cite{shu2022tia}, Cai et al.~\cite{cai2021neu}, Zhao et al.~\cite{zhao2024ret}};
                  ]
                  [.{Process}
                    \node[fill=CY,text width=6cm](t1){Shao et al.~\cite{shao2023enh}, Trivedi et al.~\cite{trivedi2023int}, Asai et al.~\cite{asai2024sel}};
                  ]
              ]
              [.{Coannotation}
                  [.{Cooperative}
                    \node[fill=CY,text width=6cm](t1){ Li et al.~\cite{Li2023Coa}, Wang et al.~\cite{Wang2024hum}
};
                  ]
                  [.{Human-guided}
                    \node[fill=CY,text width=6cm](t1){ToolCoder~\cite{zhang2023too}, Bao et al.~\cite{bao2023dis}};
                  ]
              ]
		]
       [.{Reweighting}
              [.{Sample-wise}
                 [.{Perception}
                  \node[fill=CY,text width=6cm](t1){Jiang et al.~\cite{Jiang2024imp}, Wang et al.~\cite{wang2024rew}, Thakkar et al.~\cite{thakkar2023sel}
};
                  ]
                  [.{Optimization}
                  \node[fill=CY,text width=6cm](t1){Huang et al.~\cite{huang2024div}, Chen et al.~\cite{chen2024bull}, Li et al.~\cite{li2025meta}
};
                  ]
              ]
              [.{Domain-wise}
                  [.{Experimentation}
                  \node[fill=CY,text width=6cm](t1){Abdin et al.~\cite{abdin2024phi4}};
                  ]
                  [.{Regression}
                  \node[fill=CY,text width=6cm](t1){Cai et al.~\cite{cai2024int}, Ye et al.~\cite{ye2025dat}, Liu et al.~\cite{liu2025reg}
};
                  ]
                  [.{Optimization}
                  \node[fill=CY,text width=6cm](t1){Chung et al.~\cite{chung2023uni}, Held et al.~\cite{held2025opt}, Fan et al.~\cite{fan2024dog}};
                  ]
              ]
       ]
       [.{Others}
              [.{Re-ordering}
                  [.{Training}
                  \node[fill=CY,text width=6cm](t1){Dodge et al.~\cite{dodge2020fin}, Pattnaik et al.~\cite{pattnaik2024cur}, Phi-4~\cite{abdin2024phi4}};
                  ]
                  [.{Inference}
                  \node[fill=CY,text width=6cm](t1){Xu et al.~\cite{xu2024inc}, Wu et al.~\cite{wu2024pro}, Guo et al.~\cite{guo2024makes}
};
                  ]
              ]
              [.{Prompt compression}
                  [.{Hard}
                  \node[fill=CY,text width=6cm](t1){Li et al.~\cite{li2023com}, Jiang et al.~\cite{jiang2023llm}, Chuang et al.~\cite{chuang2024lea}};
                  ]
                  [.{Soft}
                  \node[fill=CY,text width=6cm](t1){Pan et al.~\cite{pan2024llm}, Liskavets et al.~\cite{Liskavets2025pro}, Nagle et al.~\cite{Nagle2024Fun}};
                  ]
              ]
              [.{Dataset distillation}
                  \node[fill=CY,text width=7.9cm](t1){Nguyen et al.~\cite{nguyen2025syn}, Maekawa et al.~\cite{Maekawa2025}, Tao et al.~\cite{tao2024tex}, Cai et al.~\cite{cai2025kno}
};
              ]
             ] 
       ]
]
\end{scope}
\end{tikzpicture}
\caption{A taxonomy for data optimization techniques for LLMs. {\textbf{Reading guide:} The root ``Data Optimization for LLMs'' branches into four principal categories (Selection, Augmentation, Reweighting, Others), each of which further decomposes into methodological families and then into representative papers. Readers can navigate from high-level strategy to specific algorithms by following the tree from left to right.}}
\label{fig4}
\end{center}\vspace{-0.12in}
\end{figure*}

This section reviews the major data optimization techniques for LLMs within the proposed taxonomy, as illustrated in Fig.~\ref{fig4}. Rather than viewing these techniques as isolated tools, we emphasize a lifecycle perspective: different techniques become salient at different stages of LLM development, including pretraining, SFT, alignment, and inference-time optimization. From this perspective, data optimization is not only about improving individual datasets, but also about deciding \emph{when} and \emph{how} different interventions should be applied to support efficiency, effectiveness, fairness, and trustworthiness. In this sense, the techniques reviewed in this section translate earlier data-side analysis into concrete improvements in model-side objectives across the LLM lifecycle.

Under this taxonomy, existing methods are organized into four principal categories: \emph{selection}, \emph{augmentation}, \emph{reweighting}, and \emph{others}. These categories are complementary rather than mutually exclusive. In practice, data selection determines which samples are retained, augmentation expands or enriches the available data distribution, and reweighting adjusts the influence of different samples or domains during optimization. Their coordinated use often provides a more effective data pipeline than any single technique alone. To make this stage-aware view explicit, Table~\ref{tab:lifecycle_mapping} summarizes the applicability, primary roles, and typical impacts of major data optimization techniques across the LLM lifecycle.

\begin{table}[h]
\centering
\caption{Data optimization techniques across the LLM lifecycle: applicability, inherent impacts, and typical pipelines.}\vspace{-0.1in}
\label{tab:lifecycle_mapping}
\small
{
\begin{tabular}{@{}p{2.2cm}p{2.1cm}p{2.1cm}p{2.1cm}p{2.1cm}p{2.6cm}@{}}
\toprule
\textbf{Technique Category} & \textbf{Pretraining} & \textbf{SFT} & \textbf{RLHF / Alignment} & \textbf{Inference} & \textbf{Typical Pipeline} \\
\midrule
\textbf{Data Selection} & 
Heuristic filtering (language, length), quality scoring, domain reweighting, deduplication \cite{Xie2023Adv, Tirumala2023, ankner2024per} & 
Difficulty-aware selection, diversity-aware sampling, influence-based selection \cite{gao2025prin, xia2024less, li2024super} & 
Preference data selection, contrastive sampling, quality filtering for preference pairs \cite{zhou2024dav} & 
Demonstration selection for ICL, example ordering optimization \cite{guo2024makes, wu2024pro} & 
Pretraining: heuristic filter $\rightarrow$ quality score $\rightarrow$ domain reweight; SFT: difficulty-aware + diversity sampling \\
\midrule
\textbf{Data Augmentation} & 
Synthetic data generation, back-translation, paraphrasing \cite{abdin2024phi4} & 
Instruction evolution, response rewriting, counterfactual generation \cite{ding2024data, chen2023dis} & 
Preference pair augmentation, rejection sampling, self-play generation \cite{zhu2024star} & 
RAG, query rewriting \cite{Lewis2020RAG} & 
Pretraining: synthetic data for low-resource domains; SFT: instruction evolution + back-translation \\
\midrule
\textbf{Data Reweighting} & 
Domain-wise reweighting, temperature sampling, importance weighting \cite{fan2024dog, chung2023uni} & 
Sample-wise reweighting based on difficulty or value, loss adjustment \cite{Jiang2024imp, chen2024bull} & 
Preference weighting, reward modeling, importance sampling for alignment \cite{zhou2024wpo} & 
Attention reweighting, context compression weighting \cite{jiang2023llm} & 
Pretraining: domain temperature; SFT: sample difficulty weighting; RLHF: preference pair weighting \\
\midrule
\textbf{Other Techniques} & 
Dataset distillation (emerging), curriculum ordering \cite{nguyen2025syn, lee2024ITW} & 
Prompt compression for efficient fine-tuning, data merging \cite{chen2025mergeit} & 
Preference distillation, alignment tuning \cite{cai2025infialign} & 
Prompt compression, test-time data optimization \cite{jiang2023llm, li2023com, chen2025tru} & 
Pretraining: curriculum ordering; Inference: prompt compression + RAG \\
\bottomrule
\end{tabular}
}
\vspace{0.2cm}
\small{\textit{Note: ICL = In-Context Learning, RAG = Retrieval-Augmented Generation. This mapping illustrates the primary applicability of each technique; many methods can be adapted across stages with appropriate modifications.}}
\end{table}

\paragraph{Evaluation perspective.}
Evaluating data optimization methods is inherently more complex than reporting a single model-side performance number, because the effect of a data intervention depends strongly on the target stage of the LLM lifecycle, the downstream task, and the optimization objective. In pre-training, methods are often assessed through metrics such as validation perplexity, downstream transfer performance, convergence speed, training stability, and data efficiency. In SFT, evaluation more commonly relies on task accuracy, instruction-following quality, robustness, and generalization to held-out tasks. In alignment settings, studies frequently report preference win rate, reward model scores, harmlessness, and safety-related indicators. For inference-time optimization, latency, token efficiency, and response quality under constrained budgets also become important.

At the same time, benchmark usage remains highly heterogeneous across the literature. Different studies evaluate data optimization methods on different task suites, model scales, and reporting protocols, which makes direct cross-method comparison difficult. To provide a compact stage-aware summary, Table~\ref{tab:evaluation_mapping} outlines representative evaluation goals, benchmarks/tasks, and metrics commonly used for assessing data optimization methods across the LLM lifecycle, while comprehensive benchmark collections can be found in recent surveys \cite{ni2025survey,li2025systematic} and a detailed review of evaluation metrics is provided by \cite{rostam2025evaluating}.

\begin{table*}[t]
\centering
{
\caption{Representative evaluation typical goals, tasks, and metrics for assessing data optimization methods across the LLM lifecycle.}\vspace{-0.05in}
\label{tab:evaluation_mapping}
\resizebox{\textwidth}{!}{
\begin{tabular}{p{2.3cm}p{4.2cm}p{5.0cm}p{4.5cm}}
\toprule
\textbf{Lifecycle stage} & \textbf{Typical evaluation goals} & \textbf{Representative tasks} & \textbf{Common metrics} \\
\midrule
Pre-training 
& Improve corpus quality, data efficiency, transferability, and convergence behavior 
& Language modeling validation; downstream transfer evaluations; domain adaptation; multilingual and code pretraining settings 
& Perplexity; validation loss; downstream accuracy/F1; convergence speed; training stability; data efficiency \\
\midrule
Supervised fine-tuning (SFT) 
& Improve task performance, instruction-following ability, robustness, and generalization 
& Question answering, reasoning, summarization, coding, dialogue, and domain-specific instruction datasets 
& Accuracy; F1; ROUGE/BLEU; pass@k; win rate; robustness/generalization scores \\
\midrule
Alignment 
& Improve helpfulness, harmlessness, preference alignment, and safety 
& red-teaming test; safety evaluation; harmlessness and refusal test 
& Preference win rate; reward score; violation/refusal rate; safety score; human evaluation \\
\midrule
Inference-time optimization 
& Improve response quality under latency, context, or token-budget constraints 
& Prompt compression, retrieval/reranking, long-context evaluation, and test-time adaptation settings 
& Latency; token cost; memory usage; response quality; efficiency--utility trade-off \\
\bottomrule
\end{tabular}
}}
\end{table*}

Compared with the techniques summarized in \cite{wu2024data} (i.e., resampling, augmentation, perturbation, reweighting, pruning, and others), our taxonomy adopts a more consolidated view. Specifically, resampling and pruning (e.g., coreset selection) are merged into \emph{selection}, while perturbation is incorporated into \emph{augmentation}. In addition, pure data synthesis is treated separately from the main optimization taxonomy, since the primary focus of this survey is the optimization of existing data. Nevertheless, when new data are synthesized, many of the same optimization principles still apply.

A direct comparison of the relative efficacy of all technical divisions is beyond the scope of this survey. This is mainly because their performance depends strongly on the target stage, task setting, computational budget, and optimization objective. Existing studies have already shown that these methods involve non-trivial trade-offs, and no single technique is uniformly superior across all scenarios. Therefore, instead of claiming universal superiority, we focus on clarifying their applicability, representative mechanisms, and practical roles across the LLM lifecycle. Accordingly, our goal is not to rank methods globally, but to clarify the conditions under which different methods are effective, scalable, or potentially risky.

\paragraph{Failure modes and negative transfer.}
We also note that data optimization methods are not uniformly beneficial in practice. Selection may discard rare but informative samples, augmentation may introduce artifacts or semantic drift, and reweighting may overemphasize narrow objectives or domains. More broadly, improvements at one lifecycle stage or under one evaluation protocol do not always transfer to other tasks or deployment settings, and may even lead to negative transfer, reduced diversity, or safety-related regressions.

\newcommand{\xmark}{\text{\sffamily X}}
\begin{table*}[t]
\centering
\tiny
\caption{Data optimization explicitly mentioned in the technical reports of several recent foundation models.}\vspace{-0.05in}
\label{tbl2}
\renewcommand{\arraystretch}{1.5}
\begin{tabularx}{\textwidth}{| 
    >{\RaggedRight}l | 
    *{4}{>{\centering\arraybackslash\hsize=0.8\hsize}X|} 
    >{\RaggedRight\hsize=1.8\hsize}X|}
\hline
\textbf{\qquad LLMs} & \textbf{Selection} & \textbf{Augmentation} & \textbf{Reweighting} & \textbf{Re-ordering} & \textbf{\qquad Perception} \\ 
\hline\hline
Llama 3~\cite{llama3} & \checkmark & \checkmark & \xmark & \checkmark & Data quality, difficulty, diversity, redundancy, contamination \\ 
\hline
Qwen 2.5~\cite{qwen2025qwen} & \checkmark & \checkmark & \xmark & \checkmark & Data quality, diversity, redundancy, contamination \\ 
\hline
Salamandra~\cite{gonz2025sal} & \checkmark & \xmark & \xmark & \xmark & Data quality, diversity, redundancy \\ 
\hline
InternLM2~\cite{cai2024int} & \checkmark & \xmark & \checkmark & \xmark & Data quality, difficulty, redundancy, contamination \\ 
\hline
GLM~\cite{glm2024} & \checkmark & \xmark & \checkmark & \xmark & Data quality, diversity, redundancy \\ 
\hline
Phi-4~\cite{abdin2024phi4} & \checkmark & \checkmark & \checkmark & \checkmark & Data quality, difficulty, diversity, redundancy, contamination \\ 
\hline
DeepSeek-V3~\cite{deepseekai2025dee} & \checkmark & \checkmark & \xmark & \xmark & Data quality, difficulty, diversity, redundancy, balance, contamination \\ 
\hline
Hunyuan~\cite{sun2024hun} & \checkmark & \checkmark & \xmark & \xmark & Data quality, difficulty, diversity, redundancy, balance, contamination \\ 
\hline
OpenELM~\cite{mehta2024ope} & \checkmark & \xmark & \xmark & \xmark & Data quality, redundancy \\ 
\hline
Tele-FLM~\cite{li2024tel} & \checkmark & \xmark & \checkmark & \xmark & Data quality, redundancy, balance \\ 
\hline
Claude 3~\cite{claude} & \checkmark & \xmark & \xmark & \xmark & Data quality, redundancy, balance, contamination \\ \hline
{Eagle 2~\cite{li2025eagle}} & \checkmark & \checkmark & X & X & Data quality, difficulty, diversity, redundancy \\
\hline
\end{tabularx}
\end{table*}

Table~\ref{tbl2} summarizes the data optimization techniques reported for several medium and large models. {This model-oriented summary provides empirical evidence for the scalability hierarchy observed in industrial practice: data selection dominates reported pretraining pipelines (11/11 models), reflecting its feasibility for trillion-token corpora, whereas sample-wise reweighting and re-ordering appear primarily in SFT or alignment contexts, suggesting limited scalability at pretraining scale.} Some techniques, such as dataset distillation and prompt compression, are excluded because they remain rarely reported in current LLM pretraining and SFT stages. Likewise, some popular models (e.g., Gemini~\cite{geminiteam2024}) are omitted due to insufficient data-related details in their released reports.

The remainder of this section first reviews the main technical categories and then discusses how they can be integrated across stages, together with their inherent impacts and trade-offs. {Among the 280 surveyed references, 119 papers that receive substantive methodological discussion are quantitatively profiled in Table~\ref{tab:census}, revealing that data selection is the most heavily investigated path (35\%), followed by augmentation (26\%) and reweighting (16\%). While the taxonomy is organized into four principal categories, Table~\ref{tab:census} further separates re-ordering, prompt compression, and dataset distillation from 'Others' to provide a more fine-grained census.}

\begin{table}[t]
\caption{{Quantitative census of surveyed papers across major technical categories (Section 5). Only papers receiving substantive methodological discussion (at least one paragraph) are counted; peripheral citations are excluded.}}
\label{tab:census}
\centering
\small
{
\begin{tabular}{@{}lcp{6.5cm}@{}}
\toprule
\textbf{Technical Category} & \textbf{\# Papers} & \textbf{Representative Anchors} \\
\midrule
Data Selection & 42 & [2, 4, 58, 97, 121, 139, 173, 207, 238] \\
Data Augmentation & 31 & [19, 21, 29, 43, 47, 115, 182, 191, 253] \\
Data Reweighting & 19 & [27, 34, 50, 73, 78, 87, 124, 275] \\
Re-ordering & 9 & [45, 69, 112, 174, 232] \\
Prompt Compression & 11 & [14, 33, 88, 131, 138, 159, 170] \\
Dataset Distillation & 7 & [16, 114, 151, 163, 202, 212] \\
\midrule
\textbf{Total} & \textbf{119} & --- \\
\bottomrule
\end{tabular}}
\end{table}

\subsection{Data Selection}
This section organizes methods by representative approach followed by variations, highlighting differences in proxies, models, and cost–performance trade-offs. In conventional deep learning, data selection mainly focuses on either filtering noisy/erroneous samples or selecting a coreset. The first topic receives much attention as real-world data usually contain noisy/erroneous samples. However, coreset selection receives less attention as it usually employs expensive mathematical optimization schemes. In the era of LLMs, data selection nearly becomes the most important optimization technique. Both the training and inference stages are more complex than those in conventional deep learning. How to reconstruct a better dataset is of great importance in terms of both effectiveness and efficiency.

Albalak et al.~\cite{albalak2024} have conducted a systematic summary and categorization of raw data selection methodologies involved in the entire training and inference pipeline of LLMs. Wang et al.~\cite{wang2024surveyd} and Liu et al.~\cite{liu2024what} have further reviewed data selection approaches specifically pertaining to instruction tuning for LLMs, respectively. Building upon these three studies, this section will establish a novel taxonomy by integrating the data-aware dimensions discussed in the preceding section with recent advancements in synthetic data selection. This study divides previous data selection methodologies into four folds, namely heuristic rule-based, perception dimension-based, domain knowledge-based, and optimization-based. For each category, we highlight one or two representative approaches and then present other methods as variations or extensions, clarifying how they differ in terms of proxies, models, or cost-performance trade-offs.

\subsubsection{Heuristics-based}
Heuristic-based methods form the first line of defense in data optimization, leveraging simple rules or shallow classifiers for efficient, large-scale filtering. A \textbf{representative approach} in this category is the use of a lightweight, pre-trained classifier like \textbf{FastText}~\cite{joulin2016fast} for language identification and quality scoring. Its efficiency makes it a ubiquitous preprocessing step in many LLM pipelines, including those for Phi-4~\cite{abdin2024phi4} and Falcon~\cite{NEURIPS2023fa3ed726}, to rapidly remove non-target language or low-quality documents.

Building on this idea, other heuristic methods vary in their simplicity and the specific rules applied. Language filtering is an important preprocessing step for both natural and code languages; for instance, Chowdhery et al.~\cite{Chowdhery2023} employed extension-based filtering across 24 programming languages by identifying source code files matching canonical filename conventions. In the construction of Phi-4~\cite{abdin2024phi4}, a suite of customized heuristics and parsers was developed to ensure the extraction of clean, consistent training data.

Some methods rely on simple statistical metrics like text lengths or word proportions. Wang et al.~\cite{wang2025lan} filtered raw generated data by removing overly short or long samples, while Xia et al.~\cite{xia2024rethink} empirically demonstrated that length-aware data filtration serves as a stable and high-efficiency selection paradigm. Other approaches train simple classifiers or regressors based on handcrafted heuristic features; for example, Xie et al.~\cite{Xie2023Adv} trained an importance estimator parameterized by bag-of-words features, and Chang and Jia~\cite{chang2023dat} trained linear regressors to score each training example based on LLM outputs. While these methods are less sophisticated than learned approaches, their minimal computational overhead makes them indispensable for initial corpus cleaning.

\subsubsection{Perception dimension-based}
Perception dimension-based methods are utilized in most previous studies, where quality, diversity, and difficulty constitute the most extensively utilized dimensions for data selection~\cite{gao2025prin, xia2024rethink}. In contrast to heuristic-based approaches that are typically confined to the pre-processing stage, perception dimension-based methodologies are applied throughout the entire pipeline, encompassing pretraining, SFT, and inference phases. Notably, many selection methods have become fully integrated into the training process itself.

We organize these methods into two paradigms: single-dimension and multi-dimension based, with an emphasis on representative approaches while acknowledging the diversity of techniques in this space.

\textbf{Single-dimension selection.} A \textbf{representative approach} for difficulty-based selection is to use \textbf{validation loss} as a difficulty signal and prioritize samples with higher loss to avoid premature convergence during alignment~\cite{gao2025prin}. Variations of this idea employ different difficulty proxies: some use perplexity for pretraining data selection~\cite{ankner2024per} or instruction tuning~\cite{li2024super}, while others use training loss~\cite{chen2024bull} or reference model accuracy~\cite{rogoz2024unib} as alternatives. Beyond these loss- and perplexity-based methods, other studies have explored distinct paradigms for difficulty-aware selection. For instance, Zhou et al.~\cite{zhou2024dav} investigated sample selection in direct preference optimization for model alignment based on difficulty, while Vinay et al.~\cite{s2024inc} utilized influence functions to conduct demonstration selection in in-context learning. Xia et al.~\cite{xia2023moderate} proposed a ``moderate coreset'' approach that selects samples close to the score median as the final results, offering a different perspective on what makes a sample valuable. 

For diversity-based selection, a representative method involves clustering samples and then selecting from each cluster to ensure coverage~\cite{zhang2024harn, tsai2024code}. Liu et al.~\cite{liu-etal-2025-empirical} employed a density-based clustering algorithm to partition data, using question/topic counts and their variance as diversity measures. Variations may use different clustering algorithms or distance metrics~\cite{gupta2023cov}. An alternative approach to enhancing diversity is to select samples by considering their similarity with already selected samples, as demonstrated by Li and Qiu~\cite{Li2023Find} for in-context learning demonstrations. The LLM-as-a-judge paradigm has also been applied to diversity assessment: Lanchantin et al.~\cite{lanchantin2025diver} prompted an LLM to identify the most and least diverse responses within given data. Despite their implementation differences, these methods share the same objective: preserving broad coverage while avoiding concentration on highly similar samples; they mainly differ in whether diversity is operationalized through clusters, pairwise similarity, or model-based judgment.

For quality-based selection, using a trained LLM as a judge to score samples is a representative approach~\cite{li2024super, parkar2024sel}. A related line of work fine-tunes smaller language models to act as quality estimators: for example, Li et al.~\cite{li2024self} finetuned a language model to construct training examples by generating instruction prompts, and then selected high-quality examples from the generated set. Other methods for quality assessment include perturbation-based approaches, where Wang et al.~\cite{wang2024noisegpt} generated a set of perturbations for each sample and leveraged predictions to identify noisy labels. Across these methods, the key distinction lies in how quality is estimated: stronger judges often improve filtering fidelity but incur higher cost, whereas lightweight estimators offer better scalability at the expense of precision.

\textbf{Multi-dimension selection.} When more than one criterion is considered, a \textbf{representative paradigm} is \textbf{sequential filtering}, where samples are first filtered by one dimension (e.g., quality) and then by another (e.g., diversity)~\cite{pang2025imp, parkar2024sel}. For example, Parkar et al.~\cite{parkar2024sel} first used coreset-based clustering to ensure diversity, and then considered complexity to further filter the selected data. Variations include parallel aggregation, where multiple scores are combined via weighted sums or heuristics, and more sophisticated joint optimization frameworks. Liu et al.~\cite{liu2024what} examined instruction tuning data selection using complexity, quality, and diversity dimensions, where complexity scoring incorporated perplexity, instruction length, response loss, and reasoning steps.

\subsubsection{Domain knowledge-based}
When LLMs are trained for specific domains, domain knowledge can be considered in the design of data selection methods. The \textbf{foundational approach} in this category is the \textbf{Moore-Lewis criterion}~\cite{moore2010}, which uses Bayesian rule to score a sample based on its probability ratio under a general-purpose dataset and a domain-specific dataset. This idea has inspired numerous extensions: Feng et al.~\cite{feng2022aut} adapted it for encoder pretraining, while Xie et al.~\cite{Xie2023Adv} reformulated the ratio as an importance weighting for domain adaptation.

Other domain-specific methods vary in the source of domain knowledge. To avoid data contamination, Phi-4~\cite{abdin2024phi4} generated 7-gram and 13-gram features from training and test data, calculated overlap rates, and removed training samples with greater than 5\% overlap. Zhao et al.~\cite{Zhao2024clue} empirically revealed that high-quality data are more likely to have similar training dynamics to an anchor dataset, and leveraged training dynamics to select high-quality data from different domains. Ye et al.~\cite{ye2025li} hypothesized that good questions should naturally elicit extended reasoning processes, and defined three selection quantities including knowledge diversity. These approaches all stem from the same principle of leveraging domain-specific signals but differ in the particular signals used and the granularity of filtering.

\subsubsection{Optimization-based}
Optimization-based data selection methods have been extensively studied in conventional deep learning contexts, yet remain relatively scarce in the literature on LLMs. Typically, they involve defining a utility function for data subsets and then employing optimization techniques to identify the optimal subset.

A \textbf{representative approach} for utility function construction is based on \textbf{optimal transport (OT) distance}~\cite{villani2009optimal}. Kang et al.~\cite{kang2024less} provided a principled data selection method that selects training samples minimizing the OT distance between the selected set and a target distribution, directly employing state-of-the-art OT solvers~\cite{cuturi2022opt}. Variations of this idea use different distribution matching objectives. Zhang et al.~\cite{zhang2024tag} leveraged the gradient difference between the original set and a candidate subset as the utility objective. Another representative utility is directly based on learning performance: Engstrom et al.~\cite{engstrom2024ds} formulated selection as minimizing the trained model's loss on target tasks, approximated via datamodeling~\cite{ilyas2022data}. Yu et al.~\cite{yu2025data} assumed independent contributions and used a greedy selection strategy based on expected loss.

Other optimization-based frameworks have been proposed for specific scenarios. Gu et al.~\cite{gu2025data} formulated data selection for LLMs as a generalized optimal control problem and proposed an approximate optimization method to solve it. Several studies focus on sample selection in a batch setting~\cite{Wang2024great, nguyen2025mini}, which are also based on utility functions and approximate optimization. These methods share the optimization core but differ in the utility definition, the solution strategy (exact solvers, greedy approximations, optimal control, etc.), and their applicability to different stages of the LLM lifecycle. In essence, these methods differ less in their high-level goal than in how utility is defined and optimized, with more principled formulations often providing better selection quality but requiring greater computational overhead.

\subsubsection{Impact on LLM Aspects}
{Data selection exerts a multifaceted influence on LLM development. By filtering low-quality or uninformative samples, it directly enhances \textbf{trustworthiness}, reducing hallucinations and improving factual accuracy \cite{chen2023maybe, Zhang2024Ult}. Simultaneously, selection methods that prioritize diversity and balance contribute to \textbf{fairness} by mitigating representational biases \cite{li2025eagle, lu2023instag}. Recent work further demonstrates that fairness can be improved at inference time through selective prompting strategies that dynamically identify high-risk cases, avoiding the performance costs of full model retraining~\cite{ren2026whentoinvoke}. From an \textbf{effectiveness} perspective, difficulty-aware selection ensures that models encounter appropriately challenging examples, fostering complex reasoning capabilities \cite{gao2025prin, chen2024bull}. Finally, by reducing training data volume through coreset selection or pruning, data selection improves \textbf{efficiency}, enabling faster training and lower computational costs \cite{xia2024rethink, Sorscher2022}. Thus, data selection is a versatile tool that can be tailored to prioritize different LLM aspects depending on the application context.}

{While more principled selection methods often improve filtering quality, they are usually harder to scale to trillion-token corpora, where lightweight heuristics and approximate proxies remain more practical.
}

\subsection{Data Augmentation}
Data augmentation is a quite successful data optimization technique in conventional deep learning. It generates a new dataset in which samples (or features) are generated based on the raw dataset or sometimes other relevant sets~\cite{wu2024data}. Since both instance-wise and feature-wise perturbations satisfy this definition, this study incorporates perturbations into the data augmentation branch. Data augmentation has also been applied to LLMs, with extensive experimental evidence confirming its significant effectiveness~\cite{puri2023data}. Notably, retrieval-augmented generation~\cite{Lewis2020RAG} has emerged as one of the principal basic technologies for LLMs.

Wang et al.~\cite{wang2024com} recently conducted a comprehensive review of data augmentation research in deep learning. Ding et al.~\cite{ding2024data} constructed a new taxonomy for data augmentation using LLMs, which divides previous studies into four folds, namely, data creation, data labeling, data reformation, and co-annotation. Another recent survey on data augmentation for LLMs~\cite{wang2024survey} followed a similar taxonomy. Building upon these survey studies, this study divides LLMs data augmentation into four folds, namely, generation-based, perturbation-based, retrieval-based, and co-annotation-based augmentation.

\subsubsection{Generation-based augmentation}

This category of methods leverages LLMs to create more useful samples based on given samples or create labels of unlabeled samples for LLMs training or inference\footnote{{Generation-based augmentation (e.g., LLM paraphrasing) is distinct from pure synthesis: the former uses existing samples as anchors and conditions, whereas the latter generates data without such anchors. Consequently, generation-based augmentation is classified under data optimization, while pure synthesis is excluded from this taxonomy.}}.

A \textbf{representative approach} in this category is \textbf{LLM-based paraphrasing} for instruction augmentation. Zhang et al.~\cite{zhao2024imp} prompted LLMs to paraphrase original instructions into several similar ones, effectively expanding the training set. Variations of this idea include back-translation, a classical rewriting strategy extensively surveyed by Ranathunga et al.~\cite{Ranathunga2023NM} for machine translation using LLMs. Chen and Lampouras~\cite{chen2023exp} applied back-translation to augment training samples for programming language tasks. The Phi-4~\cite{abdin2024phi4} training set incorporated extensive text rewriting, with Web-rewritten content constituting 15\% of pretraining data.

Another representative approach leverages the powerful reasoning capabilities of LLMs to \textbf{annotate domain-specific data} for SFT. Zhang et al.~\cite{Zhang2024Ult} compiled the UltraMedical collections, containing instructions annotated with completions from multiple advanced LLMs. For multimodal data augmentation, Wang et al.~\cite{Wang2024Tsc} employed CoT prompting to generate high-quality CoT rationales for science question answering. As a representative implementation, Zhu et al.~\cite{zhu2024star} compiled prompt-derived ranked datasets and created pairwise comparisons for improved alignment optimization.

\subsubsection{Perturbation-based augmentation}

Different from generation-based augmentation which creates new samples from existing ones, this category of methods perturbs existing samples into different ones. Practically, perturbation-based methods exhibit lower sample alteration magnitudes than generation-based ones, though the distinction remains non-discrete. Inspired by~\cite{wu2024data}, we categorize methods by the target of perturbation.

A \textbf{representative approach} for token/word-wise perturbation is the use of classical operations such as token addition, deletion, and replacement, which have proven effective in LLM training~\cite{Lu2022Ins, 10202198}. Variations include syntactic perturbations, such as swapping subject and object in a sentence~\cite{min2020syn}, or generating small adversarial perturbations for raw texts~\cite{Zhang2024Att}. Huang et al.~\cite{Hu2024Pro} perturbed prompts by adding prefixes.

For embedding-wise perturbation, a representative method is mixup at the embedding level. Chen et al.~\cite{chen2022dou} first perturbed samples with token- or sentence-wise operations, then mixed up the embeddings of original and perturbed samples. Kim et al.~\cite{kim2023dis, kim2024ins} perturbed the logit vector of a normal sample with that of a noisy sample, demonstrating improved robustness.

For sample-wise perturbation, Yoo et al.~\cite{yoo2021gpt} applied mixup directly to raw texts and their labels. Chen et al.~\cite{chen2023dis} prompted an LLM to generate high-quality counterfactual data based on input texts. Huang et al.~\cite{huang2024enh} introduced label smoothing to enhance model alignment. Other perturbation targets include loss~\cite{wang2024llm} and reward~\cite{chen2025dis}, representing more specialized variations.

\subsubsection{Retrieval-based augmentation}

This technique is mainly used in retrieval-augmented generation (RAG) and creates augmented data from relevant data corpora~\cite{CRAG2024}. As previous surveys provide systematic reviews of RAG~\cite{zhao2024ret, gao2024ret, Fan2024ASU, li2022sur}, we focus on the retrieval component from a data optimization perspective.

A \textbf{representative approach} for similarity computation in retrieval is the use of \textbf{dense vector-based methods}, which leverage low-dimensional vectors produced by pretrained LLMs. Variations include sparse vector-based methods (e.g., BM25) suitable for keyword matching~\cite{10948482}, and task-specific methods that incorporate task characteristics such as graph structure~\cite{shu2022tia} or machine translation~\cite{cai2021neu}.

Another dimension of variation is the retrieval process itself. Gao et al.~\cite{gao2024ret} categorized processes into iterative retrieval~\cite{shao2023enh}, which repeatedly queries the knowledge base; recursive retrieval~\cite{trivedi2023int}, which refines queries through feedback loops; and adaptive retrieval~\cite{asai2024sel}, which enables LLMs to actively determine retrieval timing and content.

\subsubsection{Co-annotation-based augmentation}

This fold of methods describes human-LLM cooperative labeling systems with bidirectional refinement. We classify current research into two categories: cooperative annotation and human-guided annotation.

In \textbf{cooperative annotation}, both humans and LLMs participate. A representative approach is uncertainty-guided task allocation, where Li et al.~\cite{Li2023Coa} determined annotation task allocation between humans and LLMs by quantifying uncertainty levels of LLM-generated annotations. Wang et al.~\cite{Wang2024hum} proposed a multi-stage collaborative framework where an LLM generates annotations and explanations, then a verification model filters low-confidence samples for human review.

In \textbf{human-guided annotation}, humans compile guidelines and LLMs annotate data accordingly. Zhang et al.~\cite{zhang2023too} employed human-authored input-output pairs to guide ChatGPT in annotating tool-augmented datasets. Bao et al.~\cite{bao2023dis} formulated manual annotation guidelines for medical domains, then applied GPT-3.5 to yield AI-aligned standardized annotations.

\subsubsection{Impact on LLM Aspects}
{Data augmentation primarily enhances the \textbf{effectiveness} of LLMs by expanding the effective training distribution and improving generalization to unseen inputs \cite{ding2024data, chen2023dis}. Techniques such as counterfactual augmentation and back-translation expose models to diverse linguistic variations, strengthening robustness and reducing overfitting. Augmentation also contributes to \textbf{trustworthiness} by generating synthetic examples that cover edge cases or underrepresented scenarios, thereby improving model reliability \cite{abdin2024phi4}. However, if not carefully controlled, augmentation can inadvertently amplify biases, underscoring the need for fairness-aware augmentation strategies \cite{li2025eagle}. From an \textbf{efficiency} standpoint, augmentation can reduce the need for costly human annotation, making it a cost-effective way to scale training data \cite{parkar2024sel}. Overall, data augmentation is a powerful mechanism for improving model capabilities, provided its application is guided by an awareness of potential side effects.}

{A key limitation of augmentation-based approaches is that synthetic or perturbed samples may introduce artifacts, reduce faithfulness, or amplify pre-existing biases if quality control is insufficient.
}

\subsection{Data Reweighting}
Data reweighting constitutes one of the most prominent tools for data optimization. Data selection is actually a hard version of data reweighting: if a sample's weight is set to 0, it is not selected. We divide reweighting studies into two folds, namely sample-wise and domain-wise.

\subsubsection{Sample-wise reweighting}

The majority of reweighting studies in LLMs focus on optimizing weight allocation across different samples. Existing studies can be broadly categorized into two classes: heuristic and optimization-based.

A \textbf{representative approach} for heuristic sample reweighting is to quantify samples according to perception dimensions and then assign weights using heuristic rules. Jiang et al.~\cite{Jiang2024imp} employed a straightforward approximation method for importance weights using a pre-trained LLM and validation set. Variations of this idea use different signals for weight assignment: Wang et al.~\cite{wang2024rew} utilized reward differences between samples; Thakkar et al.~\cite{thakkar2023sel} integrated self-influence scores with pretraining representations; He et al.~\cite{he2024sof} quantified duplication degree and reduced weights of highly duplicated samples; and Zhou et al.~\cite{zhou2024wpo} reweighted preference pairs based on their probability under the current policy.

A \textbf{representative approach} for optimization-based reweighting formulates objectives that incorporate sample weights and data utilities, and then solves them via optimization. Huang et al.~\cite{huang2024div} defined a loss function on validation set and employed stochastic derivative-free optimization. Chen et al.~\cite{chen2024bull} formulated a minimax optimization problem for both model parameters and sample weights. Li et al.~\cite{li2025meta} designed a meta-learning-based method for end-to-end bi-level optimization.

\subsubsection{Domain-wise reweighting}

Domain-wise weighting is critically important for LLM training, impacting both efficiency and effectiveness. Existing methods can be categorized into three classes: experimentation-based, regression-based, and optimization-based.

A \textbf{representative approach} for experimentation-based methods leverages small-scale model experiments to derive domain weighting schemes. Abdin et al.~\cite{abdin2024phi4} observed high rank correlation between small and large model performance on data mixtures, conducting experiments on small models and transferring findings to large-scale training.

A \textbf{representative approach} for regression-based methods assumes a functional relationship between model performance and domain weights. Ye et al.~\cite{ye2025dat} established exponential regression functions relating cross-domain mixing weights to model performance. Liu et al.~\cite{liu2025reg} proposed a LightGBM-based regression framework~\cite{Ke2017lig} with superior experimental outcomes.

A \textbf{representative approach} for optimization-based methods directly formulates optimization objectives. Chung et al.~\cite{chung2023uni} defined a simple objective with constraints for domain proportions. Held et al.~\cite{held2025opt} added a utility-based term. Fan et al.~\cite{fan2024dog} trained a proxy small model to pursue domain weights via bi-level optimization. Pan et al.~\cite{pan2024sca} proposed a scalable bi-level optimization algorithm for models with over 20B parameters.

\subsubsection{Impact on LLM Aspects}
{Data reweighting offers fine-grained control over the learning process, enabling targeted improvements across multiple LLM aspects. For \textbf{effectiveness}, reweighting based on difficulty or value ensures that challenging or high-impact samples receive appropriate emphasis, accelerating convergence and improving final performance \cite{Jiang2024imp, chen2024bull}. In terms of \textbf{fairness}, reweighting can directly address representational imbalances by upweighting underrepresented groups or downweighting overrepresented ones \cite{fan2024dog, chung2023uni}. For \textbf{trustworthiness}, reweighting can mitigate the influence of noisy or poisoned samples, enhancing model robustness and reliability \cite{zhou2024wpo}. Finally, domain-wise reweighting during pretraining shapes the model's knowledge distribution, impacting both \textbf{effectiveness} (by balancing general and specialized knowledge) and \textbf{efficiency} (by focusing compute on high-value domains) \cite{ye2025dat}. Data reweighting thus serves as a unified mechanism for aligning model behavior with diverse optimization objectives.} 

{Although reweighting provides finer control than hard selection, poorly calibrated weights may destabilize optimization or overemphasize narrow signals at the expense of broader generalization.}

\subsection{Other Strategies}
Beyond the three categories previously discussed, numerous additional optimization methodologies have been investigated. Due to space limitations, we briefly introduce three representative techniques.

\subsubsection{Data re-ordering}

LLM performance demonstrates notable sensitivity to data ordering~\cite{feng2024max}, manifesting in two dimensions: (1) sequential arrangement of training samples, and (2) positional ordering of demonstrations during inference.

{A \textbf{representative approach} for training sample ordering is curriculum learning. Pattnaik et al.~\cite{pattnaik2024cur} explored preference learning with easier-to-harder preference pairs. In Phi-4 pretraining~\cite{abdin2024phi4}, training tokens were allocated to filtered Web data first, followed by synthetic token-dominated mixtures. Many models order training by sample length, progressing from shorter to longer samples~\cite{qwen2025qwen}.}

For in-context learning demonstration ordering, Xu et al.~\cite{xu2024inc} revealed that different orderings can cause performance to vary from near random to near SOTA, formulating example ordering as an optimization problem. Wu et al.~\cite{wu2024pro} employed pretrained language model embeddings with neural bandit optimization. Guo et al.~\cite{guo2024makes} filtered orderings prone to inducing prediction imbalances, then selected the ordering with maximum influence.

\subsubsection{Prompt compression}

Prompt compression is a parameter-efficient technique that compresses complex prompts while preserving semantic fidelity. Following Li et al.~\cite{li2024pro}, we categorize methods into hard prompt and soft prompt approaches.

A \textbf{representative approach} for hard prompt compression prunes redundant tokens. Li et al.~\cite{li2023com} enhanced inference efficiency by identifying and deleting redundancy in input context. Jiang et al.~\cite{jiang2023llm} designed a coarse-to-fine compression strategy integrating budget control and token-level dependency modeling.

A \textbf{representative approach} for soft prompt compression uses encoder-decoder architectures. Pan et al.~\cite{pan2024llm} framed compression as token classification ensuring faithfulness. Liskavets et al.~\cite{Liskavets2025pro} proposed sentence-level context-aware compression. Nagle et al.~\cite{Nagle2024Fun} provided a rate-distortion framework unifying token-level methods.

\subsubsection{Dataset distillation}

This technique distills large datasets into small synthetic sets such that models trained on them achieve comparable generalization~\cite{10273632}. While mainly investigated in computer vision, NLP applications are emerging.

{A \textbf{representative approach} for text dataset distillation uses gradient matching. Nguyen et al.~\cite{nguyen2025syn} iteratively optimized synthetic example embeddings to match target gradients, then transformed them to text tokens with low perplexity. Variations include Maekawa et al.~\cite{Maekawa2025} utilizing gradient-based optimization for few-shot NLP tasks, Tao et al.~\cite{tao2024tex} leveraging LLM embeddings for data-efficient generation, and Cai et al.~\cite{cai2025kno} automating distillation from scientific literature.}

\subsubsection{Impact on LLM Aspects}
{The techniques grouped under ``others'' also have distinct impacts on LLM aspects. Data re-ordering, particularly in curriculum learning and in-context example ordering, directly influences \textbf{effectiveness} by shaping the sequence in which models encounter information, often leading to better generalization and reasoning \cite{guo2024makes, wu2024pro}. Prompt compression primarily targets \textbf{efficiency}, reducing token consumption during inference without sacrificing output quality \cite{jiang2023llm, li2023com}. Dataset distillation, though still emerging for LLMs, holds promise for improving \textbf{efficiency} by compressing large corpora into compact synthetic sets, while also offering potential benefits for \textbf{trustworthiness} by enabling controlled data generation \cite{nguyen2025syn, tao2024tex}. Collectively, these techniques complement the main optimization categories and address specialized needs across the LLM lifecycle.}

{These techniques are often highly effective in specialized scenarios, but their gains can be sensitive to task structure, prompt format, or inference constraints, which limits their universality.
}


\subsection{Integrating Data Optimization Across the LLM Lifecycle}\label{sec5.6}

{The techniques surveyed above are not mutually exclusive; rather, they can and should be orchestrated across the LLM development lifecycle to achieve compound benefits. Table~\ref{tab:lifecycle_mapping} maps each optimization category to the stages where it is most applicable, offering a stage-aware perspective. Typical pipelines compose these methods sequentially: selection first establishes a clean subset, augmentation expands coverage, and reweighting adjusts learning emphasis—the output of one stage becoming the input to the next, with per-stage objectives shifting from efficiency (pretraining) to precision (SFT) to alignment (RLHF). Conceptually, selection defines the model's \textbf{knowledge boundary}, augmentation expands its \textbf{coverage}, and reweighting calibrates \textbf{learning emphasis} within that boundary. Below, we discuss how these techniques complement each other throughout the lifecycle and articulate their inherent impacts on model development.}

\paragraph{Pretraining.}
Pretraining establishes the foundational knowledge of LLMs and operates at the largest scale, making efficiency a primary concern. Data selection dominates this stage, beginning with coarse-grained heuristic filtering (e.g., language identification, length filtering, deduplication) to establish a clean baseline corpus \cite{Xie2023Adv, Tirumala2023}. More sophisticated selection methods, such as quality scoring based on perplexity or classifier-based filtering, further refine the corpus \cite{abdin2024phi4, ankner2024per}. Domain-wise reweighting, often guided by small-scale experiments or optimization frameworks, shapes the final data mixture to balance general capabilities with domain-specific performance \cite{fan2024dog, ye2025dat}. Recent frameworks like CLIMB demonstrate how clustering-based selection and iterative optimization can automatically discover optimal pretraining mixtures without manual domain labels \cite{climb2025}. Data augmentation plays a comparatively minor role at this stage due to scale, though synthetic data is increasingly used to supplement under-represented domains \cite{abdin2024phi4}. The inherent impact of pretraining data optimization is twofold: it determines both the upper bound of model capabilities and the efficiency with which those capabilities are acquired.

\paragraph{Supervised Fine-Tuning (SFT)}
SFT adapts pretrained models to follow instructions and perform specific tasks. At this stage, data volume is smaller, enabling more computationally intensive techniques. Data selection focuses on identifying high-quality instruction-response pairs, often using difficulty metrics \cite{gao2025prin}, diversity-aware sampling \cite{xia2024less}, or influence-based methods \cite{xia2024less} to curate compact yet effective training sets. Remarkably, frameworks like InfiAlign demonstrate that principled data selection can achieve state-of-the-art reasoning performance using only 12\% of the training data \cite{cai2025infialign}. Data augmentation becomes highly effective at this stage, with techniques such as instruction evolution, response rewriting, and counterfactual generation expanding the effective training distribution \cite{ding2024data, chen2023dis}. Sample-wise reweighting modulates the contribution of individual examples based on difficulty or value, enabling fine-grained control over learning priorities \cite{Jiang2024imp, chen2024bull}. The inherent impact of SFT data optimization is targeted capability enhancement: well-curated data can elicit complex behaviors (e.g., chain-of-thought reasoning) that pretraining alone does not fully develop.

\paragraph{RLHF and Alignment.}
Alignment stages, including RLHF and direct preference optimization, aim to align model behavior with human preferences and values. Data selection here focuses on curating high-quality preference pairs, often using quality filtering or contrastive sampling to identify informative comparisons \cite{zhou2024dav}. Data augmentation through rejection sampling or self-play generation expands preference datasets without expensive human annotation \cite{zhu2024star}. Reweighting techniques adjust the influence of preference pairs based on their discriminative value or alignment with desired behaviors \cite{zhou2024wpo}. The inherent impact of alignment data optimization is value-directed steering: it shapes not just what models can do, but how they choose to behave.

\paragraph{Inference.}
At inference time, data optimization focuses on the input context rather than the training corpus. Demonstration selection for in-context learning identifies the most informative examples for few-shot tasks, with ordering optimization further refining performance \cite{guo2024makes, wu2024pro}. RAG dynamically augments inputs with relevant external knowledge \cite{Lewis2020RAG}. Prompt compression techniques reduce input length while preserving semantic content, improving efficiency without sacrificing quality \cite{jiang2023llm, li2023com}. Emerging work on test-time data optimization suggests that dynamic refinement of inputs during inference can yield substantial performance gains, particularly for reasoning tasks \cite{chen2025tru}. The inherent impact of inference-time optimization is adaptive performance: it enables models to leverage computation adaptively based on task demands.

\paragraph{Cross-Stage Integration.}
Beyond stage-specific applications, optimization techniques can interact across stages. For instance, insights from inference-time demonstration selection can inform SFT data curation by identifying examples that are most effective for in-context learning. Similarly, value estimation methods developed for data selection can be repurposed for sample reweighting in alignment stages. The Chameleon framework exemplifies this cross-stage potential, providing a unified approach to domain reweighting that transfers seamlessly across pretraining and finetuning \cite{chameleon2025}. As LLM development becomes increasingly multi-stage and iterative, such integrative perspectives on data optimization will become essential.

While these techniques can be composed across the LLM lifecycle, it is crucial to recognize that they operate under fundamental theoretical constraints. We discuss these inherent limitations in Section~\ref{sec:fundamental-limitations}.

\begin{table}[t]
\centering
\caption{Objective-driven data optimization: typical technique matches and trade-offs.}
\label{tab:objective_mapping}
\small
\setlength{\tabcolsep}{3pt}
\begin{tabular}{@{}p{2.8cm}p{3cm}p{8cm}}
\toprule
\textbf{Optimization Objective} & \textbf{Typical Techniques} & \textbf{Rationale, Typical Scenarios, and Trade-offs} \\
\midrule
\textbf{Accuracy First} & 
Quality-focused data selection \cite{chen2023maybe, Zhang2024Ult}; 
Domain-specific augmentation \cite{ding2024data, chen2023dis}; 
Difficulty-aware reweighting \cite{gao2025prin, chen2024bull} & 
Prioritizes data fidelity and task alignment over computational efficiency. Suitable for high-stakes domains like medical diagnosis, legal analysis, and scientific reasoning where error tolerance is low. 
\textbf{Trade-off}: May increase computational cost; \textbf{if token-constrained}: prioritize quality selection over augmentation to stay within budget. \\
\midrule
\textbf{Efficiency First} & 
Lightweight selection (heuristic, perplexity-based) \cite{xia2024rethink, ankner2024per}; 
Dataset distillation \cite{nguyen2025syn, tao2024tex}; 
Coreset selection \cite{Sorscher2022, xia2023moderate} & 
Aims to reduce training data volume while preserving model performance. Ideal for rapid prototyping, resource-constrained environments, or when training multiple model variants. 
\textbf{Trade-off}: May sacrifice accuracy; \textbf{if accuracy non-negotiable}: use difficulty-aware selection instead of random pruning. \\
\midrule
\textbf{Cost Saving} & 
Data reweighting \cite{Jiang2024imp, chen2024bull}; 
Active learning \cite{parkar2024sel}; 
Influence-based selection \cite{xia2024less, kwon2024data} & 
Maximizes performance gains from limited annotation or acquisition budgets. Suitable for scenarios with expensive human labeling (e.g., medical annotation) or restricted access to third-party data. 
\textbf{Trade-off}: May underexplore rare but important cases if value estimation is imperfect; \textbf{if coverage is critical}: combine with diversity-aware selection or targeted augmentation. 
 \\
\midrule
\textbf{Knowledge Coverage} & 
Diversity-aware sampling \cite{li2025eagle, lu2023instag}; 
Retrieval-based augmentation \cite{Lewis2020RAG}; 
Domain-aware reweighting \cite{fan2024dog, ye2025dat} & 
Ensures broad representation across languages, domains, or tasks. Critical for multilingual models, domain adaptation, and general-purpose assistants. 
\textbf{Trade-off}: Broader coverage may dilute specialization or introduce noisier samples; \textbf{if downstream precision is critical}: pair with quality filtering or domain-aware selection. 
\\
\midrule
\textbf{Token Saving} & 
Prompt compression \cite{jiang2023llm, li2023com}; 
Demonstration selection \cite{guo2024makes, xu2024inc}; 
Context pruning \cite{jiang2023llm} & 
Reduces input length without substantially sacrificing output quality. Essential for inference optimization, long-context tasks, and cost-sensitive API usage (pay-per-token). 
\textbf{Trade-off}: Aggressive compression may remove task-relevant context; \textbf{if accuracy is sensitive to context}: prefer selective demonstration pruning over heavy prompt compression. 
 \\
\midrule
\textbf{Balanced / Multi-Objective} & 
Hybrid pipelines (e.g., RAG + LoRA) \cite{Lewis2020RAG, Hu2024Pro}; 
Pareto-aware optimization \cite{chameleon2025} & 
Combines techniques to navigate trade-offs between competing objectives. Increasingly adopted in production systems where no single objective dominates. 
\textbf{Trade-off}: Hybrid pipelines are more flexible but also harder to tune and evaluate; \textbf{practical use}: they require multi-metric validation rather than a single performance score.
 \\
\bottomrule
\end{tabular}
\end{table}

\subsection{Usage-Oriented Perspectives: Matching Techniques to Objectives}

The previous sections have surveyed data optimization techniques from a methodological perspective. However, practitioners often face a different question: given a specific optimization objective—such as maximizing accuracy, minimizing cost, or improving efficiency—which techniques are most commonly associated with success? Table~\ref{tab:objective_mapping} offers an integrated view of these tendencies, mapping objectives to the techniques typically leveraged in the literature, along with the accompanying rationales and trade-offs. This framework is intended to be generative rather than definitive, helping researchers navigate the landscape by highlighting common patterns.

\paragraph{Accuracy-First Scenarios.}
In high-stakes applications such as medical diagnosis, legal analysis, or scientific reasoning, model accuracy is paramount and often outweighs considerations of computational efficiency or cost. For these scenarios, quality-focused data selection methods that filter out noisy or low-quality samples are essential \cite{chen2023maybe, Zhang2024Ult}. However, filtering-based strategies should be applied with care, as overly aggressive criteria may remove rare but useful samples and reduce data diversity. Domain-specific augmentation, such as generating synthetic medical cases or legal precedents, can further enhance task-specific performance \cite{ding2024data, chen2023dis}. Difficulty-aware reweighting ensures that challenging examples receive appropriate emphasis during training, preventing models from overfitting to easy patterns \cite{gao2025prin, chen2024bull}. The key trade-off is that these methods often require more computational resources and careful validation, but the performance gains justify the investment in critical applications.

\paragraph{Efficiency-First Scenarios.}
When rapid prototyping, resource constraints, or large-scale experimentation are primary concerns, efficiency becomes the dominant objective. Lightweight selection methods based on heuristics or perplexity can dramatically reduce training data volume with minimal computational overhead \cite{xia2024rethink, ankner2024per}. Dataset distillation offers a more sophisticated approach, compressing entire datasets into compact synthetic sets that retain essential information \cite{nguyen2025syn, tao2024tex}. Coreset selection methods identify representative subsets that approximate full-data performance \cite{Sorscher2022, xia2023moderate}. The trade-off is that aggressive pruning may sacrifice some performance, but the efficiency gains often enable broader experimentation and faster iteration cycles.

\paragraph{Cost-Saving Scenarios.}
When annotation budgets are limited or data acquisition is expensive, maximizing performance per unit cost becomes the primary objective. Data reweighting techniques adjust the influence of samples based on their estimated value, allowing limited data to have maximum impact \cite{Jiang2024imp, chen2024bull}. Active learning frameworks iteratively select the most informative samples for annotation, reducing the total annotation burden \cite{parkar2024sel}. Influence-based methods identify samples that contribute most to model performance, enabling targeted data acquisition \cite{xia2024less, kwon2024data}. The key insight is that not all samples are equally valuable; strategic selection can achieve strong performance with a fraction of the annotation cost.

\paragraph{Knowledge Coverage Scenarios.}
For applications requiring broad generalization across languages, domains, or tasks—such as multilingual models or general-purpose assistants—ensuring adequate knowledge coverage is critical. Diversity-aware sampling methods explicitly promote representation across different categories, preventing models from overfitting to dominant patterns \cite{li2025eagle, lu2023instag}. Retrieval-based augmentation dynamically incorporates external knowledge during inference, expanding effective coverage without retraining \cite{Lewis2020RAG}. Domain-aware reweighting balances contributions from different data sources to prevent under-representation of niche areas \cite{fan2024dog, ye2025dat}. The trade-off is that maximizing coverage may dilute performance on any single task, but the resulting models are more robust and adaptable.

\paragraph{Token-Saving Scenarios.}
With the rise of pay-per-token API pricing and the challenges of long-context processing, reducing token usage without sacrificing output quality has become an important objective. Prompt compression techniques identify and remove redundant tokens while preserving semantic content \cite{jiang2023llm, li2023com}. Demonstration selection for in-context learning chooses the most informative examples, reducing the number of tokens required for few-shot prompting \cite{guo2024makes, xu2024inc}. Context pruning methods remove irrelevant portions of long documents before feeding them to the model \cite{jiang2023llm}. These techniques are particularly valuable for inference optimization and cost-sensitive deployments.

\paragraph{Balanced and Multi-Objective Scenarios.}
In real-world production systems, multiple objectives often compete for priority. A financial chatbot, for example, may need both accuracy (for regulatory compliance) and token efficiency (for cost management). In such cases, hybrid pipelines that combine techniques are increasingly adopted. For instance, RAG provides broad knowledge coverage with low training cost, while LoRA fine-tuning injects domain-specific expertise with minimal overhead \cite{Lewis2020RAG, Hu2024Pro}. More sophisticated approaches explicitly model trade-offs between objectives~\cite{qian2025demandclean, chameleon2025}. As the field matures, such multi-objective optimization is likely to become the norm rather than the exception.

\paragraph{Practical Takeaways.}
For practitioners navigating this landscape, several guiding principles emerge: (1) Start by clearly defining your primary optimization objective—different goals demand different techniques. (2) Consider the full lifecycle: techniques suitable for pretraining may not transfer directly to inference optimization. (3) Embrace hybrid approaches: combining complementary techniques often yields better results than pursuing a single method to its extreme. (4) Evaluate holistically: a technique that improves accuracy at the cost of massive token consumption may not be optimal in production settings. (5) {Ground decisions in theoretical insights:} recent theoretical advances can provide actionable guidance for practical design. For instance, \citet{rezaei2026highdimensional} prove that covariance matching is the optimal strategy for synthetic data selection in high-dimensional settings, suggesting that practitioners should prioritize matching second-order statistics between synthetic and real data distributions rather than relying solely on heuristic quality filters. By matching techniques to objectives with these principles in mind, practitioners can navigate the rich landscape of data optimization methods more effectively.

\begin{table}[t]
\centering
\caption{Scenario-based mapping: when to apply different data optimization techniques.}
\label{tab:scenario_mapping}
\small 
\setlength{\tabcolsep}{3pt} 
\begin{tabular}{@{}p{1.9cm}p{1.9cm}p{1.9cm}p{1.9cm}p{1.9cm}p{1.9cm}p{1.9cm}@{}}
\toprule
\textbf{Technique} & \textbf{Pre} & \textbf{IT} & \textbf{Align} & \textbf{Infer} & \textbf{Data-Ltd} & \textbf{Compute-Ltd} \\
\midrule
\textbf{Selection} & 
High \newline (filter, dedup, mix) \cite{Xie2023Adv, Tirumala2023, abdin2024phi4} & 
High \newline (diff, div, influence) \cite{gao2025prin, xia2024less, li2024super} & 
Med \newline (pref filter) \cite{zhou2024dav} & 
High \newline (demo select) \cite{guo2024makes, wu2024pro} & 
High \newline (high-value) \cite{xia2024less, kwon2024data} & 
High \newline (coreset) \cite{Sorscher2022, xia2023moderate} \\
\midrule
\textbf{Augmentation} & 
Low \newline (synthetic) \cite{abdin2024phi4} & 
High \newline (evolve, para) \cite{ding2024data, chen2023dis} & 
Med \newline (reject sampling) \cite{zhu2024star} & 
High \newline (RAG) \cite{Lewis2020RAG} & 
High \newline (generate) \cite{parkar2024sel} & 
Low \newline (costly) \\
\midrule
\textbf{Reweighting} & 
High \newline (domain mix) \cite{fan2024dog, chung2023uni} & 
Med \newline (diff weight) \cite{Jiang2024imp, chen2024bull} & 
High \newline (pref weight) \cite{zhou2024wpo} & 
Low & 
High \newline (max impact) \cite{Jiang2024imp} & 
Low \\
\midrule
\textbf{Other} & 
Med \newline (distill) \cite{nguyen2025syn} & 
Med \newline (prompt comp) \cite{chen2025mergeit} & 
Low & 
High \newline (prompt comp) \cite{jiang2023llm, li2023com} & 
Med \newline (distill) \cite{nguyen2025syn} & 
High \newline (distill, comp) \cite{nguyen2025syn, jiang2023llm} \\
\bottomrule
\end{tabular}

\vspace{0.2cm}
\footnotesize{\textit{Note: High/Med/Low indicate typical practical relevance rather than universal superiority. Pre=Pretraining, IT=Instruction Tuning, Align=Alignment (RLHF), Infer=Inference, Data-Ltd=Data-Limited, Compute-Ltd=Compute-Limited. ICL=In-Context Learning, RAG=Retrieval-Augmented Generation.}}
\end{table}

{These techniques are not applied in isolation. In practice, practitioners can narrow down suitable methods through a simple usage-oriented decision process: first identify the \textbf{lifecycle stage} (e.g., pretraining, instruction tuning, alignment, or inference), then determine the dominant \textbf{practical constraint} (e.g., data scarcity or limited compute), and finally choose the optimization category that best matches the scenario (Table~\ref{tab:scenario_mapping}). For example, pretraining under compute constraints typically favors lightweight selection and domain reweighting, whereas instruction tuning under data-limited settings often benefits more from augmentation and value-aware reweighting. Typical pipelines then compose these techniques sequentially: selection first establishes a clean subset, augmentation expands coverage, and reweighting adjusts learning emphasis. The output of one stage becomes the input to the next, with per-stage objectives shifting from efficiency (pretraining) to precision (SFT) to alignment (RLHF).}

\subsection{{Fundamental Limitations of Data Optimization}}
\label{sec:fundamental-limitations}

While the techniques surveyed in this section offer practical improvements for LLM development, they operate under fundamental theoretical constraints that impose inherent limits on their efficacy. Understanding these limitations is essential for setting realistic expectations and guiding principled method selection.

\subsubsection{No Free Lunch in Data Optimization}

The \textbf{No Free Lunch (NFL) theorems} for machine learning establish that no single learning algorithm can universally outperform all others across every possible data distribution and task configuration~\cite{wolpert1997no}. Applied to data optimization, this implies that any strategy optimized for specific data properties (e.g., high-quality web text) must necessarily underperform on alternative distributions (e.g., low-resource languages or specialized domains). Consequently, the ``optimal'' technique is always \textbf{task-relative}: a method that excels for general-domain pretraining may fail for vertical applications, and vice versa.

\subsubsection{Information-Theoretic Lower Bounds}

Data compression techniques face strict information-theoretic limits. For $(\epsilon,k,z)$-clustering problems, coreset construction requires at least ${\Omega(k/\epsilon^2)}$ points in Euclidean space to preserve solution costs within $(1\pm\epsilon)$ factor~\cite{cohen2022towards}---a lower bound that is tight and cannot be circumvented by algorithmic innovation. For panel data regression, any coreset must retain observations from {all $N$ individuals} to guarantee approximation quality under certain parameter configurations~\cite{huang2020coresets}. These results establish that aggressive data reduction inevitably sacrifices information fidelity, with the loss bounded below by fundamental geometric and statistical constraints.

\subsubsection{Associated Side Effects}

Every major optimization category entails inherent side effects that persist even with advanced methodological refinements, 
necessitating conscious trade-off decisions

\paragraph{Data Selection.} 
Selection necessarily discards information; even samples deemed ``low-quality'' 
may contain rare but valuable signals for tail-domain generalization. 
The Information Bottleneck principle indicates that aggressive pruning without 
proper guidance increases distortion and can exceed finite-sample bounds, 
degrading generalization~\cite{tishby2015deep}. 
{Empirically, Gao et al.~\cite{gao2025prin} demonstrated that overly 
difficult preference examples significantly degrade alignment performance 
across multiple LLMs and datasets, decreasing win rates by up to 9.4\% 
when such examples are included in DPO training on UltraFeedback-binarized data, while Dai et al.~\cite{dai2025improving} 
showed that influence-based selection suffers from selection length bias and decreasing effectiveness over the training trajectory.}

\paragraph{Data Augmentation.} 
Augmentation inevitably introduces distribution shift. 
Synthetic or perturbed samples may occupy regions of the input space that 
diverge from the true data manifold, creating ``hallucinated'' patterns that 
impair robustness. 
{Havrilla et al.~\cite{havrilla2024survey} systematically evaluated 
quality-diversity-complexity trade-offs in synthetic data, showing that 
over-optimization of one dimension (e.g., quality) at the expense of others 
can degrade out-of-distribution generalization, potentially undermining the 
benefits of data augmentation.} 
Counterfactual augmentation, while improving coverage, can induce a 
phenomenon where models focus exclusively on edited causal 
features and exclude other non-edited causal features, limiting 
out-of-distribution generalization~\cite{fan2024unlock}.

\paragraph{Data Reweighting.} 
Reweighting modifies the empirical risk landscape; poorly calibrated weights 
can destabilize optimization dynamics or overemphasize narrow subpopulations, 
leading to forgetting of broader knowledge. 
{Chung et al.~\cite{chung2023uni} demonstrated that temperature-based 
sampling with excessive upsampling of low-resource languages in multilingual 
pretraining leads to overfitting, memorization, and inefficient use of 
training compute, particularly as model scale increases.} 
In fairness-oriented applications, this tension is especially visible: 
reducing disparity for one sensitive attribute may come at the cost of 
lower overall accuracy~\cite{zafar2019fairness}, or may inadvertently 
exacerbate disparities along untargeted attributes under specific 
contexts~\cite{paraschou2026intra}. Poorly calibrated weights may 
also amplify noise in minority subgroups and conflict with privacy 
objectives~\cite{qian-etal-2025-tug}.

\subsubsection{The Multi-Objective Negotiation}

The practical implication is that data optimization is inherently a \textbf{trade-off negotiation} rather than a universal improvement. Practitioners must accept that:
\begin{itemize}
    \item optimizing for efficiency (e.g., aggressive pruning) will, with mathematical certainty, sacrifice some accuracy;
    \item optimizing for fairness on one attribute may degrade performance on others;
    \item optimizing for synthetic data coverage may introduce distribution artifacts.
\end{itemize}

{\textbf{Fairness constraints are structurally distinct}. 
Unlike accuracy-efficiency trade-offs that are often amenable to 
Pareto optimization, 
fairness optimization faces \textbf{unavoidable} limitations: 
\textbf{neuron superposition} creates zero-sum competition with 
privacy~\cite{qian-etal-2025-tug}; \textbf{contextual fragility} prevents 
stable generalization across deployment settings~\cite{karvonen2025robustly}; 
and \textbf{metric misalignment} precludes reliable 
validation~\cite{li2024survey}.}

The optimal strategy depends on the specific alignment between the technique's inductive biases and the target task's characteristics. Rather than pursuing universally optimal solutions, effective data optimization requires \textbf{deliberate trade-off management} informed by theoretical limits and empirical validation across multiple objectives.

\section{Connections and Future Directions}\label{sec6}
\subsection{Connections}
Four aspects, namely, perception, technical paths, similarity/opposition, and cooperation, connect different methods within a technical path or across different paths.

\subsubsection{Connections via perceptions}
Data perception (explicit/implicit) constitutes the foundational stage within data optimization workflows, where methodologies spanning divergent technical paths in Section~\ref{sec5} may converge on shared quantitative metrics, thereby establishing connections across heterogeneous paths. For example, many methods from different techniques rely on quality and difficulty in data selection~\cite{sharma2024tex}~\cite{thrush2025imp}~\cite{tian2025dee}, augmentation~\cite{jia2025cur}~\cite{yang2023boo}, and reweighting~\cite{held2025opt}~\cite{sow2025dyn}. Other quantities including uncertainty and value are also used in different optimization techniques.

The use of similar perception quantities indicates shared underlying principles or theoretical foundations across these methods. Notably, these perception dimensions and optimization approaches have been extensively explored in conventional deep learning, potentially offering valuable insights and inspiring novel research directions for LLMs.

\begin{figure*}[t] 
    \centering \includegraphics[width=0.98\linewidth,height=1.96in]{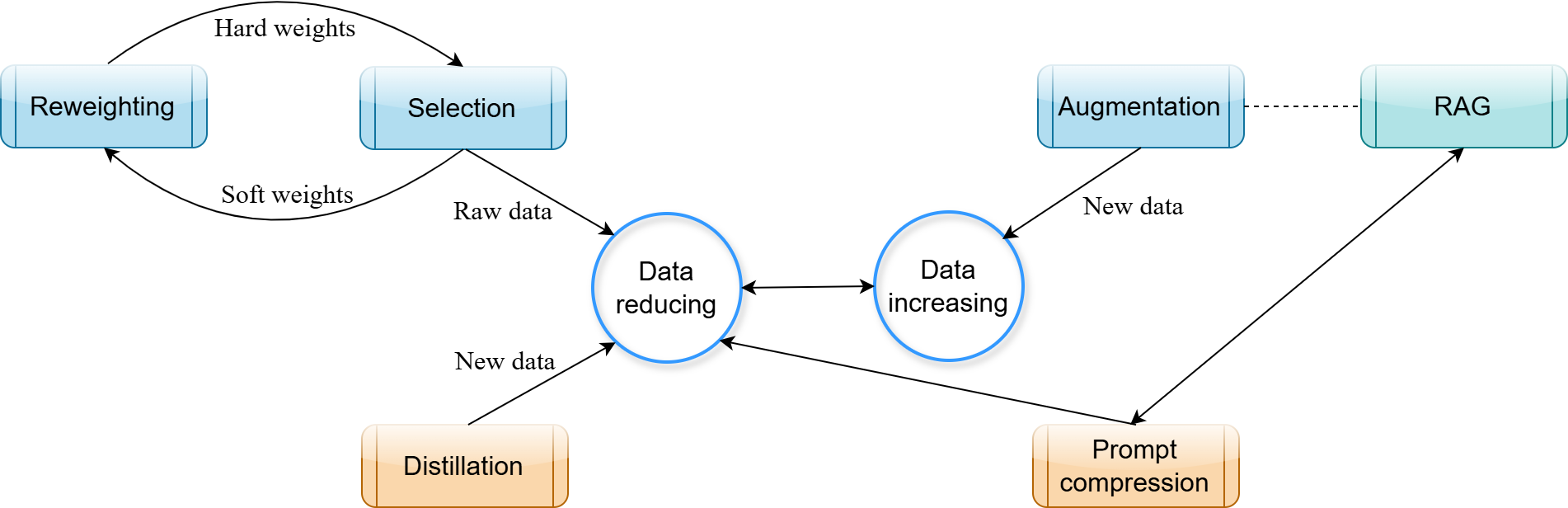}    \vspace{-0.0in}
    \caption{Similarity/opposition between the main data optimization techniques.}
    \label{connect}
     \vspace{-0.08in}
\end{figure*}
\subsubsection{Connections via strategies}
At least three strategies are widely used in different technical paths, namely, mathematical optimization, LLM-based, and small model-based.

In the context of data selection and reweighting, mathematical optimization serves as a fundamental methodology. This involves formulating objective functions tailored to instance selection (via hard weights) or instance weighting (via soft weights), followed by optimization procedures to derive optimal weight allocations.

The utilization of SOTA LLMs, which demonstrate exceptional generation and reasoning capabilities, has emerged as a key approach for scoring, ranking, selection, or augmentation across various tasks and applications~\cite{li2025gen}. As illustrated in Fig.~\ref{fig2}, this methodology has been successfully implemented across three primary technical pathways and three additional auxiliary pathways, demonstrating promising performance. Its application scope is expected to expand significantly in future research and applications.

Given the high computational demands of LLMs, many algorithms now optimize data selection by leveraging training dynamics or prediction performance from smaller-scale models. This strategy has proven broadly effective in practice, which is used in data selection~\cite{ankner2024per}~\cite{yang2024sma}, reweighting, and distillation.

\subsubsection{Connections via similarity/opposition}
The similar and opposite relationships existing among the technical paths are introduced in Section~\ref{sec5}.

As shown in Fig.~\ref{connect}, data selection and reweighting are closely related techniques, as their key steps are nearly the same. If the weights are taken as sampling rate or binarization, sample reweighting becomes selection. Although data selection and dataset distillation are closely related techniques aiming to obtain compact, high-quality datasets, they diverge in optimization priorities. While both enhance training/inference efficiency, data selection additionally seeks to improve model performance, whereas dataset distillation primarily focuses on maintaining acceptable performance levels despite reduced data scale. In addition, selection outputs raw data, whereas distillation outputs new data.

While data selection/distillation and augmentation represent contrasting operational paradigms, they demonstrate paradoxical alignment in pursuing shared optimization objectives for learning systems. As illustrated in Fig.~\ref{connect}, these techniques exhibit non-trivial methodological convergence through their implementation mechanisms. It is believable that more intrinsic connections can be established for them. Specifically, prompt compression and RAG represent two contrasting directions. The former aims to minimize prompts, while the latter seeks to enrich contextual information. Nevertheless, there are studies that integrate these two contradictory strategies into a more effective one~\cite{he2025cod}.

\subsubsection{Connections via composition}
The technical pathways presented in Section~\ref{sec5} can be systematically integrated to establish a more comprehensive and effective data optimization framework by combining two or more of the discussed optimization methods.

Augmentation adds controlled variety by making changes like swapping words or adjusting sentence structures. Data selection can remove less useful or harmful augmented instructions~\cite{li2024emp}. Therefore, one intuitive composition strategy is to combine data augmentation with data filtering or reweighting. Huang et al.~\cite{huang2024div} segmented long demonstrations into shorter segments (which can be regarded as a patch-like augmentation method), then fused the logits of different segments through a reweighting approach. This approach not only improved storage efficiency but also enhanced accuracy.
Kuo et al.~\cite{kuo2025llm} designed two efficient, automatic weighted-loss approaches to closely align the distribution of synthetic augmented data with that of real-world data. Jiang et al.~\cite{Jiang2024imp} integrated data augmentation, selection, and reweighting for LLMs self-improvement.

Other possible compositions can be explored in future work. For example, dataset distillation can be performed after filtering noisy or erroneous samples. For instance, Tukan et al.~\cite{tukan2023dat} merged the idea of data subset selection with dataset distillation by training the distilled set on important samples. Reweighting distilled samples for training may achieve more satisfactory performance. For example, Cui et al.~\cite{cui2024mit} employed a sample reweighting scheme to reduce bias amplification in dataset distillation.

\subsection{Future Directions}
The techniques surveyed above have matured, yet several structural gaps persist.  
We consolidate the future landscape into five directions, merging overlapping themes from our original eight.  
Within each direction, we highlight concrete technical desiderata that the community can act upon.

\begin{itemize}

\item[(1)] \textbf{Theoretical frameworks, scalable valuation, and reproducible benchmarking.} 
While this survey has systematically categorized data optimization techniques and identified their individual limitations, the field currently lacks unified theoretical frameworks to rigorously compare methods across the full design space. 
For instance, the decision between hard selection and soft reweighting involves nuanced trade-offs between granularity of control, computational overhead, and optimization stability---yet these dimensions are rarely studied in combination. 
We call for future research to develop: 
(a) formal models that characterize the \emph{applicability conditions} of different techniques (e.g., under what data distributions does perplexity-based selection provably underperform?); 
(b) standardized benchmarking protocols that control for confounding variables (model scale, training duration, evaluation tasks) to enable fair cross-method comparison; and 
(c) \emph{negative results} repositories to overcome publication biases and document failure modes systematically. 
A specific, underexplored algorithmic challenge within this space is \textbf{scalable data valuation}: current Shapley-based and influence-based methods are computationally prohibitive for trillion-token corpora, calling for sub-linear approximation algorithms (e.g., sketching-based or sampling-based estimators) that correlate with ground-truth influence without full model retraining. 
Cross-pollination from tabular data-centric AI---where reinforcement-learning-based data transformation and automated data-repair agents have been systematically studied~\cite{wang2026toward}---offers methodological templates that can be adapted to LLM training corpora, particularly for optimizing tabular metadata (document provenance, quality scores, temporal stamps) jointly with raw text. 
Such frameworks would transform survey-level observations into actionable, theory-grounded decision tools.

\item[(2)] \textbf{Explainable data optimization}. 
Explainable AI has emerged as an important research direction in the past few years~\cite{9050829}. 
Enhancing the explainability of data optimization processes remains a critical challenge. 
The optimization algorithms can achieve better performance if the runners of the algorithms understand thoroughly both the procedures and the results of the algorithms. 
There have been studies on interpretable data reweighting~\cite{Zhou2024Inv} and augmentation~\cite{10.1145/3533048}.

\item[(3)] \textbf{Trustworthy, safety-aligned, and multimodal data optimization}. 
Current LLMs encounter trustworthy and safety-critical challenges such as hallucinations and adversarial vulnerabilities. 
Addressing these requires dual-path optimization: preventive filtering using retrieval-augmented metrics to downweight unverifiable claims, and corrective augmentation injecting counterfactual examples with verification signals. 
Dataset distillation could prioritize factually dense samples (e.g., scientific abstracts over fictional narratives) while curriculum learning sequences data by verifiability. 
The filtering process in pretraining could further combine toxicity classifiers with ethical conflict detection to suppress subtly harmful examples, such as gendered stereotypes masked in natural language, and adversarial augmentation might generate safety-critical edge cases (e.g., prompts that elicit harmful outputs) to strengthen alignment during training. 
There are some studies focusing on trustworthy issues in data optimization techniques, such as trustworthy dataset distillation~\cite{Ma2025Tow} and trustworthy data creation~\cite{Liang2022Adv}, as well as studies exploring safety-aligned topics~\cite{hui2025tox,kumar2024eth,10.1145/3691620.3695018}. 
Extending these safeguards to vision--language and speech--text settings remains largely open: cross-modal analogues of quality, diversity, and redundancy metrics must be defined to operate at billion-pair scale, and modality-aware reweighting is needed to address modal imbalance where text tokens dominate visual tokens. Concrete instantiations include cross-modal data selection~\cite{wu2026scalselect} and vision-text synthetic data generation for augmentation, both governed by the same perceptual dimensions. While this survey focuses on text-based LLMs due to the maturity of the literature, the taxonomy in Figs.~\ref{fig2}, \ref{fig3}, and~\ref{fig4} is architecturally agnostic to modality; the primary adaptation required is in the perception metrics, not in the optimization pipeline itself. 

\item[(4)] \textbf{Autonomous and test-time data optimization}. 
The AI agent has emerged as a rapidly advancing automation technology in recent years. 
While automated training, or more specifically, the training agents~\cite{zeng2025aut}, has become a prominent research area for LLMs, research on applying agent technologies to data optimization has received comparatively limited attention. 
Ooi et al.~\cite{Oi2024} systematically reviewed research on AI-powered data systems and proposed a novel high-level architecture for AI-powered autonomous data systems. 
This architecture actually enables automated data optimization applicable to LLMs. 
It achieves adaptation by integrating LLM-based data perception modules (e.g., LLM-as-a-judge\footnote{We also note that LLM-as-a-judge, while scalable and convenient, may be sensitive to the evaluator model, prompt design, and scoring criteria, and may not always align with human judgment. It should therefore be treated as an auxiliary signal rather than a fully reliable substitute for human evaluation.}) into the data analysis stage. 
Zhou et al.~\cite{Zhou2024sta} established an automated pipeline to enhance the data quality through multi-agent collaboration and assessment. 
Kang et al.~\cite{kang2025aut} designed a two-stage, scale-aware data composition framework for domain reweighting. 
Luo et al.~\cite{luo2026dataagentslevelsstate} proposed a classification system for data agents modeled after autonomous driving levels, providing guidelines for the design of data agents. 
Complementing these training-time agents, test-time refinement offers untapped potential: recent studies demonstrate that test-time scaling can further improve the performance of LLMs, enabling significant improvements in both specialized and general reasoning tasks~\cite{zhang2025wha}. 
Techniques could include dynamic prompt compression. 
Cotta and Maddison~\cite{cotta2024tes} explored robust test-time strategies to control biases for LLMs. 
Chen et al.~\cite{chen2025tru} presented a cost-efficient strategy to improve LLM performance by scaling test-time compute based on repeated data sampling.

\item[(5)] \textbf{Full-lifecycle and multi-objective data optimization}. 
Current data optimization primarily focuses on training and inference stages. 
In practice, LLMs require continuous updating, including but not limited to few-shot SFT and knowledge unlearning, while simultaneously requiring mitigation of acquired knowledge forgetting. 
For example, Zhao et al.~\cite{zhao2026rethinking} revealed that online reweighting offers better generalization than offline. 
This persistent update process inherently involves data-related considerations, thereby evidently presenting optimization potential within data utilization frameworks. 
As discussed in Section 3.1, the four optimization goals may exhibit complex trade-offs rather than pure synergies. 
Current data optimization research typically focuses on optimizing for a single objective (e.g., selecting data to maximize accuracy, or reweighting data to enhance fairness). 
A promising future direction is to develop methods that can explicitly navigate these trade-offs, allowing practitioners to specify preferences or constraints across multiple objectives. 
This may involve formulating data optimization as a multi-objective optimization problem, developing Pareto-optimal data strategies, or creating evaluation protocols that holistically measure a method's impact across all four dimensions.
\end{itemize}

\section{Conclusions}\label{sec7}
This paper has attempted to summarize a wide range of data-centric studies within an independent deep learning realm, namely, data optimization for LLMs. A taxonomy for data optimization, as well as a taxonomy for data perception, is established for previous technical paths on data selection, augmentation, reweighting, and others developed for LLMs. Connections among different technical paths are discussed, and potential future directions are presented. {Throughout this survey, we have highlighted how these optimization techniques collectively enhance LLM effectiveness, trustworthiness, fairness, and efficiency—demonstrating that data-centric interventions are fundamental to advancing model capabilities.} Our study provides a purely systematic data-centric view for the development of LLMs. In the future, we will explore a more comprehensive taxonomy on data optimization incorporating studies for both conventional deep learning and LLMs.

\bibliographystyle{ACM-Reference-Format}
\bibliography{ReferenceFormat}

\end{document}